\documentclass[journal]{IEEEtran}
\usepackage{amsmath,amssymb,amsthm}
\usepackage{graphicx}
\usepackage{booktabs}
\usepackage{multirow}
\usepackage{array}
\usepackage{algorithm}
\usepackage{algpseudocode}
\usepackage{cite}
\usepackage{xcolor}
\usepackage{tikz}
\usetikzlibrary{arrows.meta,positioning,fit,calc}
\usepackage[hidelinks]{hyperref}

\newtheorem{proposition}{Proposition}

\theoremstyle{remark}

\graphicspath{{figures/}}
\newcommand{\nCaDEMax}{6}
\newcommand{\nCaDEMin}{1.7}
\newcommand{\nCaGaVsGaMax}{112}
\newcommand{\nCaGaVsGaMin}{12}
\newcommand{\nCaMultiSucc}{110}
\newcommand{\nCaPSOMax}{17}
\newcommand{\nCaPSOMin}{2.3}
\newcommand{\nCaVsLsbMultiMax}{2.3}
\newcommand{\nCaVsLsbMultiMin}{1.3}
\newcommand{\nGaSolvedMax}{80}
\newcommand{\nGaSolvedMin}{0}
\newcommand{\nLsbMultiSucc}{96}
\newcommand{\nMcnemarP}{0.001}
\newcommand{\nMcnemarTopoP}{0.727}
\newcommand{\nMedTimeCa}{203}
\newcommand{\nMedTimeTopo}{383}
\newcommand{\nNdOZAckleyCa}{100}
\newcommand{\nNdOZAckleyLsb}{70}
\newcommand{\nNdOZRosenbrockCa}{100}
\newcommand{\nNdOZRosenbrockLsb}{95}
\newcommand{\nNdVAckleyCa}{80}
\newcommand{\nNdVAckleyLsb}{90}
\newcommand{\nNdVRosenbrockCa}{100}
\newcommand{\nNdVRosenbrockLsb}{90}
\newcommand{\nNlsMultiCa}{6.7}
\newcommand{\nNlsMultiLsb}{1.4}
\newcommand{\nNlsMultiTopo}{7.8}
\newcommand{\nOracleQuadMaxPct}{0.04}
\newcommand{\nOracleTopMaxPct}{0.09}
\newcommand{\nRatioMultiDELsbMax}{2.44}
\newcommand{\nRatioMultiDELsbMin}{0.67}
\newcommand{\nRatioMultiIsoMax}{1.00}
\newcommand{\nRatioMultiIsoMin}{0.64}
\newcommand{\nRatioMultiPSOLsbMax}{3.47}
\newcommand{\nRatioMultiPSOLsbMin}{1.04}
\newcommand{\nRatioMultiSurfMax}{1.11}
\newcommand{\nRatioMultiSurfMin}{0.99}
\newcommand{\nRatioMultiTopoMax}{1.55}
\newcommand{\nRatioMultiTopoMin}{1.15}
\newcommand{\nSuccMultiDECa}{110}
\newcommand{\nSuccMultiDELsb}{103}
\newcommand{\nSuccMultiIso}{113}
\newcommand{\nSuccMultiPSOCa}{108}
\newcommand{\nSuccMultiPSOLsb}{94}
\newcommand{\nSuccMultiSurf}{110}
\newcommand{\nSuccMultiTopo}{108}
\newcommand{\nWilcoxTimeTopoP}{1.3\times10^{-3}}

\newcommand{\nMonoFits}{6094}
\newcommand{\nMonoBad}{93}
\newcommand{\nMonoBadPct}{1.5}

\newcommand{\nRayVolTS}{0.0026}

\newcommand{\nvAFiveDYW}{37}
\newcommand{\nvAFiveDYN}{48}
\newcommand{\nvATenDYW}{46}
\newcommand{\nvKTenLowDYWTL}{16/11/3}
\newcommand{\nvKTenLowESAWTL}{1/23/6}
\newcommand{\nvKTenLowTSWTL}{10/20/0}
\newcommand{\nvKThirtyLowDYWTL}{3/17/0}
\newcommand{\nvKThirtyLowESAWTL}{0/19/1}
\newcommand{\nvKThirtyLowTSWTL}{6/12/2}
\newcommand{\nvALowDYW}{83}
\newcommand{\nvALowDYT}{13}
\newcommand{\nvALowDYL}{0}
\newcommand{\nvALowDYN}{96}
\newcommand{\nvALowESAW}{1}
\newcommand{\nvALowESAT}{37}
\newcommand{\nvALowESAL}{58}
\newcommand{\nvALowESAN}{96}
\newcommand{\nvALowTSW}{64}
\newcommand{\nvALowTST}{28}
\newcommand{\nvALowTSL}{4}
\newcommand{\nvBTwentyDYWTL}{0/2/0}
\newcommand{\nvBTwentyESAWTL}{1/1/0}
\newcommand{\nvBTwentyTSWTL}{2/0/0}
\newcommand{\nvATwentyDYW}{11}
\newcommand{\nvATwentyDYL}{0}
\newcommand{\nvATwentyDYN}{12}
\newcommand{\nvATwentyESAWTL}{1/2/9}
\newcommand{\nvATwentyTSWTL}{8/2/2}
\newcommand{\nvAllTwentyDYN}{17}
\newcommand{\nvBThirtyDYWTL}{0/0/2}
\newcommand{\nvBThirtyESAWTL}{1/0/1}
\newcommand{\nvBThirtyTSWTL}{2/0/0}
\newcommand{\nvAThirtyDYW}{10}
\newcommand{\nvAThirtyDYL}{0}
\newcommand{\nvAThirtyESAWTL}{0/2/10}
\newcommand{\nvAThirtyTSWTL}{4/6/2}
\newcommand{\nvKTenHighDYWTL}{3/3/0}
\newcommand{\nvKTenHighESAWTL}{0/2/4}
\newcommand{\nvKTenHighTSWTL}{2/4/0}
\newcommand{\nvAHighDYW}{21}
\newcommand{\nvAHighDYN}{24}
\newcommand{\nvAHighDYWTL}{21/3/0}
\newcommand{\nvAHighESAL}{19}
\newcommand{\nvAHighESAN}{24}
\newcommand{\nvAHighTSWTL}{12/8/4}
\newcommand{\nvAblAllFiveBestWTL}{28/48/0}
\newcommand{\nvAblAllTenBestWTL}{56/20/0}
\newcommand{\nvAblALowExtrapWTL}{1/92/3}
\newcommand{\nvAblALowBestWTL}{58/38/0}
\newcommand{\nvAblAllLowExtrapW}{3}
\newcommand{\nvAblAllLowExtrapT}{145}
\newcommand{\nvAblAllLowExtrapL}{4}
\newcommand{\nvAblAllLowExtrapN}{152}
\newcommand{\nvAblAllLowBestW}{84}
\newcommand{\nvAblAllLowBestT}{68}
\newcommand{\nvAblAllLowBestL}{0}
\newcommand{\nvAblAllLowBestN}{152}
\newcommand{\nvKeepKTenLowOneWTL}{8/22/0}
\newcommand{\nvKeepAllLowOneW}{18}
\newcommand{\nvKeepAllLowOneL}{1}
\newcommand{\nvKeepAllLowOneN}{152}
\newcommand{\nvKeepAllHighOneW}{4}
\newcommand{\nvKeepAllHighOneL}{0}
\newcommand{\nvKeepAllHighOneN}{34}
\newcommand{\nvKeepKOneLowOneWTL}{0/30/0}
\newcommand{\nvKeepPFewLowOneWTL}{5/54/1}

\newcommand{\nvTimeCASAEATen}{5.9}
\newcommand{\nvTimeDYTen}{1.0}
\newcommand{\nvTimeESATen}{2.1}
\newcommand{\nvTimeTSTen}{1.1}
\newcommand{\nvMapsTen}{23}
\newcommand{\nvGlobTen}{23}
\newcommand{\nvGlobMinTen}{20}
\newcommand{\nvGlobMaxTen}{61}
\newcommand{\nvTimeCASAEAThirty}{74}
\newcommand{\nvTimeDYThirty}{45}
\newcommand{\nvTimeESAThirty}{22}
\newcommand{\nvTimeTSThirty}{11}
\newcommand{\nvGlobThirty}{23}
\newcommand{\nvLeadFromFive}{10}
\newcommand{\nvLeadToFive}{25}
\newcommand{\nvMRankCASAEAFive}{2.1}
\newcommand{\nvMRankDYFive}{3.6}
\newcommand{\nvMRankESAFive}{1.2}
\newcommand{\nvMRankTSFive}{3.0}
\newcommand{\nvLeadFromTen}{10}
\newcommand{\nvLeadToTen}{15}
\newcommand{\nvMRankCASAEATen}{2.0}
\newcommand{\nvMRankDYTen}{3.8}
\newcommand{\nvMRankESATen}{1.3}
\newcommand{\nvMRankTSTen}{2.9}
\newcommand{\nvMRankCASAEATwenty}{1.8}
\newcommand{\nvMRankDYTwenty}{3.7}
\newcommand{\nvMRankESATwenty}{1.6}
\newcommand{\nvMRankTSTwenty}{3.0}
\newcommand{\nvMRankCASAEAThirty}{2.2}
\newcommand{\nvMRankDYThirty}{3.7}
\newcommand{\nvMRankESAThirty}{1.2}
\newcommand{\nvMRankTSThirty}{2.9}
\newcommand{\nvPkFiftyPFewFiveCASAEA}{90}
\newcommand{\nvPkTenPFewFiveOne}{73}
\newcommand{\nvPkFiftyPFewFiveOne}{77}
\newcommand{\nvPkFiftyKTenFiveCASAEA}{52}
\newcommand{\nvPkFiftyKTenFiveOne}{40}
\newcommand{\nvPkFiftyPFewTenCASAEA}{87}
\newcommand{\nvPkTenPFewTenOne}{74}
\newcommand{\nvPkFiftyPFewTenOne}{75}
\newcommand{\nvPkFiftyKTenTenCASAEA}{65}
\newcommand{\nvPkFiftyKTenTenOne}{46}
\newcommand{\nvPkFiftyPFewTwentyCASAEA}{85}
\newcommand{\nvPkFiftyPFewTwentyOne}{74}
\newcommand{\nvPkFiftyKTenTwentyCASAEA}{85}
\newcommand{\nvPkFiftyKTenTwentyOne}{68}
\newcommand{\nvPkFiftyPFewThirtyCASAEA}{80}
\newcommand{\nvPkFiftyPFewThirtyOne}{76}
\newcommand{\nvPkFiftyKTenThirtyCASAEA}{48}
\newcommand{\nvPkFiftyKTenThirtyOne}{33}
\newcommand{\nvPkFiftyPOppMinLow}{81}
\newcommand{\nvPkFiftyPOppMaxLow}{97}
\newcommand{\nvPkFiftyPOppMinHigh}{71}
\newcommand{\nvPkFiftyPOppMaxHigh}{83}

\newcommand{\ovPkTenPFewTenCASAEA}{79}
\newcommand{\ovPkFiftyPFewTenCASAEA}{80}
\newcommand{\ovPkFiftyPFewTenDY}{87}
\newcommand{\ovPkFiftyPFewTenESA}{87}
\newcommand{\ovPkFiftyPFewTenTS}{94}

\newcommand{\SuppRef}[1]{\ifcsname suppnum@#1\endcsname\csname suppnum@#1\endcsname\else\textbf{??}\fi}
\expandafter\def\csname suppnum@s:example\endcsname{S-I}
\expandafter\def\csname suppnum@s:oraclend\endcsname{S-II}
\expandafter\def\csname suppnum@tab:oraclend\endcsname{SI}
\expandafter\def\csname suppnum@s:nd\endcsname{S-III}
\expandafter\def\csname suppnum@s:cons\endcsname{S-IV}
\expandafter\def\csname suppnum@fig:gp\endcsname{S1}
\expandafter\def\csname suppnum@fig:nd\endcsname{S2}
\expandafter\def\csname suppnum@tab:cons\endcsname{SII}
\expandafter\def\csname suppnum@s:conv2d\endcsname{S-V}
\expandafter\def\csname suppnum@s:dev\endcsname{S-VI}
\expandafter\def\csname suppnum@tab:sens\endcsname{SIII}
\expandafter\def\csname suppnum@tab:params\endcsname{SIV}
\expandafter\def\csname suppnum@s:params\endcsname{S-VII}
\expandafter\def\csname suppnum@s:oneshot\endcsname{S-VIII}
\expandafter\def\csname suppnum@s:eand\endcsname{S-IX}
\expandafter\def\csname suppnum@fig:conv2d\endcsname{S3}
\expandafter\def\csname suppnum@tab:oneshot\endcsname{SV}
\expandafter\def\csname suppnum@s:ablation\endcsname{S-X}
\expandafter\def\csname suppnum@s:height\endcsname{S-XI}
\expandafter\def\csname suppnum@tab:eand\endcsname{SVI}
\expandafter\def\csname suppnum@tab:ablA\endcsname{SVII}
\expandafter\def\csname suppnum@tab:ablother\endcsname{SVIII}
\expandafter\def\csname suppnum@s:impl\endcsname{S-XII}
\expandafter\def\csname suppnum@tab:height\endcsname{SIX}
\expandafter\def\csname suppnum@s:proofs\endcsname{S-XIII}
\expandafter\def\csname suppnum@s:repro\endcsname{S-XIV}
\expandafter\def\csname suppnum@s:default\endcsname{S-XV}
\expandafter\def\csname suppnum@tab:defdev\endcsname{SX}
\expandafter\def\csname suppnum@tab:defdim\endcsname{SXI}
\expandafter\def\csname suppnum@s:raystress\endcsname{S-XVI}
\expandafter\def\csname suppnum@sec:supp-recent\endcsname{S-XVII}
\expandafter\def\csname suppnum@tab:raystress\endcsname{SXII}
\expandafter\def\csname suppnum@fig:mqex\endcsname{S4}
\expandafter\def\csname suppnum@s:mq\endcsname{S-XVIII}
\expandafter\def\csname suppnum@tab:init\endcsname{SXIII}
\expandafter\def\csname suppnum@tab:portcheck\endcsname{SXIV}
\expandafter\def\csname suppnum@tab:zhen\endcsname{SXV}
\expandafter\def\csname suppnum@tab:budget\endcsname{SXVI}
\expandafter\def\csname suppnum@s:onepeak\endcsname{S-XIX}
\expandafter\def\csname suppnum@tab:peak\endcsname{SXVII}
\expandafter\def\csname suppnum@tab:mqA\endcsname{SXVIII}
\expandafter\def\csname suppnum@tab:mqother\endcsname{SXIX}
\expandafter\def\csname suppnum@tab:mqhigh\endcsname{SXX}
\expandafter\def\csname suppnum@tab:oldmq\endcsname{SXXI}
\expandafter\def\csname suppnum@tab:oldpeak\endcsname{SXXII}
\expandafter\def\csname suppnum@tab:olddim\endcsname{SXXIII}
\expandafter\def\csname suppnum@tab:oldpeakhigh\endcsname{SXXIV}
\expandafter\def\csname suppnum@tab:dev\endcsname{SXXV}

\newcommand{\thFunc}{Func.}

\newcommand{\thShape}{Peak shape}
\newcommand{\thBand}{band}

\newcommand{\thNls}{LS per run}
\newcommand{\shQuad}{quadratic}
\newcommand{\shGauss}{Gaussian}
\newcommand{\shSkew}{skewed Gaussian}
\newcommand{\shCone}{cone}
\newcommand{\shPeaks}{Peaks (main peak)}

\newcommand{\R}{\mathbb{R}}
\newcommand{\bx}{\mathbf{x}}
\newcommand{\bu}{\mathbf{u}}
\newcommand{\bg}{\mathbf{g}}
\newcommand{\bH}{\mathbf{H}}
\newcommand{\bS}{\boldsymbol{\Sigma}}

\AtBeginDocument{}

\begin{document}

\title{CA-EC: Contour-Assisted Evolutionary\\Computation With Peak Structure Learning}

\author{Jun Zhang,~\IEEEmembership{Fellow,~IEEE}, Tong Qian, Hua Ding, Wei-Jie Yu, Zhi-Hui Zhan,~\IEEEmembership{Fellow,~IEEE}, and Jian-Yu Li%
\thanks{This work was supported in part by the National Research Foundation of Korea under Grants RS-2025-00555463 and RS-2025-25456394. \emph{(Corresponding author: Jun Zhang.)}}%
\thanks{Jun Zhang is affiliated with Hanyang University ERICA, Republic of Korea; Nankai University; and Chaoyang University of Technology, Taiwan (e-mail: junzhang@ieee.org).}%
\thanks{Tong Qian, Hua Ding, Zhi-Hui Zhan, and Jian-Yu Li are with Nankai University, Tianjin, China (e-mail: qiantongnku@163.com; dinghua\_ieee@163.com; zhanapollo@163.com; lijianyu@hanyang.ac.kr).}%
\thanks{Wei-Jie Yu is with Sun Yat-sen University, Guangzhou, China (e-mail: yuweijie6@mail.sysu.edu.cn).}}
\markboth{}{}

\maketitle

\begin{abstract}
The solutions evaluated during the run of an evolutionary algorithm (EA) carry information about the peak structure of the fitness landscape. This paper proposes contour-assisted evolutionary computation (CA-EC), which estimates the locations, heights and shapes of the peaks from the contours (level sets) of a surface interpolating the evaluated solutions and directs the search to unexplored peaks. Peaks are identified by a persistence-filtered merge tree of superlevel sets in two dimensions and by hill climbing with ray-cast level sets in higher dimensions. Around a smooth peak in $d$ dimensions, the volume of the superlevel set raised to the power $2/d$ vanishes linearly at the peak height and its centroid approaches the peak location linearly; fitting these laws to nested level sets extrapolates peak height and location, and second moments give the shape. CA-EA inserts contour steps into a genetic algorithm, differential evolution or particle swarm optimization, each starting local search from the extrapolated location of a peak selected by extrapolated height. CA-SAEA, for expensive optimization, enters each peak at the surface maximum and samples it with the estimated shape within its own trust radius; since maps built from few solutions rank peaks unreliably, it maintains several candidate peaks and progressively reduces them to one. On two-dimensional benchmarks, CA-EA requires substantially fewer evaluations than the underlying EAs; with $50d$ evaluations, CA-SAEA reaches lower errors than DYCORS and TS-DDEO on most functions in five to twenty dimensions, and than DYCORS in thirty.
\end{abstract}

\begin{IEEEkeywords}
Contour tree, evolutionary computation, expensive optimization, landscape analysis.
\end{IEEEkeywords}

\section{Introduction}
\label{sec:intro}

\IEEEPARstart{A}{n} evolutionary algorithm (EA) explores a search space by evaluating a stream of candidate solutions. After a few generations it has typically evaluated hundreds of points, many of them on the slopes of the most promising regions. Selection uses these evaluations to rank individuals, and variation operators use the ranks to generate new ones; the spatial information carried by the evaluated points, which is in effect a sparse survey of the landscape, is largely left unused. Learning-aided evolution~\cite{zhan2023leo} trains a neural network on successful parent--offspring pairs to move individuals towards better positions, and knowledge learning evolutionary computation~\cite{jiang2025klec} trains a network that maps the position of an individual to a successful evolutionary direction, using the improvements collected in every generation; both learn how to improve individuals rather than where the peaks of the landscape lie. Surrogate-assisted evolutionary algorithms (SAEAs)~\cite{jin2011surrogate,jin2019datadriven,li2022ecexpensive} exploit this information by fitting a model to the evaluated points and by using the model to pre-screen candidates~\cite{liu2014gpeme,sun2017sacoso,wang2017calsapso,yu2018shpso,wei2021llso} or to choose infill points~\cite{jones1998ego,gutmann2001rbf,regis2007srbf,regis2013dycors}. In these methods the surrogate is used as a point predictor: it answers ``how good is this candidate?'', but it is rarely asked ``how many peaks does the landscape have, where are they, and how high might each of them be?''

Surveyors face the second question routinely. From a set of spot heights they draw contour lines, and a peak (a summit, in surveying terms) shows up as a family of nested closed contours. Even when no spot height lies on the peak itself, the shrinking contours indicate where the peak is and, by the rate at which the enclosed area shrinks with altitude, how high it rises. This paper asks whether an EA can read its own evaluated solutions in the same way: draw the contour map implied by the solutions evaluated so far, identify the peaks it contains, estimate the location, height and shape of each peak from its contour lines, and spend further evaluations accordingly. A first step in this direction was the contour method of~\cite{lin2008contour}, later adapted to niching in ANDE~\cite{wang2020ande}: around each locally best point of the population it extrapolates linearly from a few neighbors to one contour level slightly above that point's value, and it evaluates the centroid of the extrapolated points as a guess of where the optimum lies. We use instead the contour map of an interpolated surface: it decomposes the evaluated region into peaks, and the nested contours of every peak give an estimate of its location, height and shape.

Turning this picture into an algorithm raises five difficulties. First, contour lines are only trustworthy where there are data; outside the region covered by the evaluated points an interpolated surface is extrapolation and produces spurious peaks. Second, an interpolated surface has ripples, so genuine peaks must be distinguished from artifacts. Third, a quantitative rule is needed to infer a peak from its contours. Fourth, beyond two dimensions contour lines become level sets that cannot be traced on a grid. Fifth, the estimates must be turned into search decisions without wasting evaluations, which matters most when evaluations are expensive.

We call the resulting framework contour-assisted evolutionary computation (CA-EC) and address these difficulties as follows. The surface is reconstructed with radial basis functions on log-compressed values, and it is contoured only inside a \emph{mapped area} defined by the data. A \emph{contour tree} (merge tree) of the superlevel sets~\cite{carr2003contourtree}, pruned by topological \emph{persistence}~\cite{edelsbrunner2002persistence,edelsbrunner2010book,dey2022ctda}, identifies the significant peaks. For every peak, the \emph{data-supported band} of nested closed contours gives, through an area law, an estimate of the location, height and shape of the peak. In $n$ dimensions the contours become level sets, which we measure by casting rays from the peak of the interpolated surface. Two algorithms use the map: a contour-assisted EA (CA-EA) that accelerates a genetic algorithm (GA), differential evolution (DE) or particle swarm optimization (PSO) in the middle of a run, and a contour-assisted SAEA (CA-SAEA) in which every expensive evaluation is chosen with the help of the surrogate and the contour map.

The contributions of this paper are as follows.
\begin{enumerate}
\item The CA-EC framework, in which the search learns the peak structure of the landscape from the solutions it has already evaluated. This is the main contribution of the paper. The evaluated region is decomposed into peaks, a memory records the peaks already explored, and evaluations are allocated among the peaks. In two dimensions the decomposition combines a mapped area with a persistence-pruned contour tree; in $n$ dimensions the peaks are located by hill climbing on the interpolated surface and described by ray-cast level sets.
\item Two algorithms. CA-EA adds to GA, DE and PSO a contour step with a memory of the explored peaks and a Nelder--Mead search started from an extrapolated peak. CA-SAEA allocates expensive evaluations between surrogate-screened offspring and sampling around the peaks of the contour map, which it enters at the maximum of the interpolated surface, with a trust radius and a memory for every peak.
\item An experimental study from two to thirty dimensions. CA-EA is tested on two-dimensional, five-dimensional and ten-dimensional benchmarks. CA-SAEA is tested with $50d$ evaluations on multi-peak landscapes of our own design, whose peaks differ in height and orientation and are elongated to different degrees (condition numbers up to $10^6$), and on classic functions, in five to thirty dimensions, against DYCORS~\cite{regis2013dycors}, ESA~\cite{zhen2023esa} and TS-DDEO~\cite{zhen2023twostage}. We state which landscapes the method is built for, and we report in the same way the functions on which it has no advantage. Three components of CA-SAEA were revised during its development, which we disclose. The last of them, a first phase that keeps several peaks, was added after first experiments on these landscapes; all results reported on them are therefore from instances on which no algorithm had been run before the algorithm was fixed (Section~\ref{sec:exp-setup}).
\item Implementations of all methods, the generator of the test landscapes, all experiment scripts and all raw results, which will be released publicly upon publication. The EAs, CA-EA and its main controls, DYCORS and the initial design of CA-SAEA have two independent implementations (Section~\ref{sec:exp-setup}).
\end{enumerate}

The experiments lead to three conclusions. First, in two dimensions CA-EA reduces the number of evaluations required by a GA by factors of \nCaGaVsGaMin\ to \nCaGaVsGaMax; most of this gain comes from the local search itself, and on the functions with many local optima the GA with contour steps is \nCaVsLsbMultiMin\ to \nCaVsLsbMultiMax\ times faster than the GA with local search from its best point (Section~\ref{sec:exp-caea}). Second, on the landscapes the method is built for, those with one to five smooth peaks and the classic functions dominated by one peak, CA-SAEA is better than the well-established DYCORS on \nvALowDYW\ of \nvALowDYN\ functions in five and ten dimensions and worse on none, and better than the recent TS-DDEO on \nvALowTSW\ and worse on \nvALowTSL; in twenty and thirty dimensions the counts are \nvAHighDYWTL\ against DYCORS and \nvAHighTSWTL\ against TS-DDEO (Sections~\ref{sec:exp-mq} and~\ref{sec:exp-dim}). Third, a decision taken from an early map should not be final: a version of CA-SAEA that follows the peak ranked highest by its first map stays on that peak, and keeping several peaks first raises the share of the ten-dimensional runs that end on the highest of three or five peaks from \nvPkFiftyPFewTenOne\% to \nvPkFiftyPFewTenCASAEA\%, the level that the other three algorithms reach. The limits of the method are reported in the same way: the recent ESA reaches a higher accuracy on the smooth peaks, and the advantage over DYCORS and TS-DDEO becomes smaller as the number of peaks grows. A contour map is by nature a low-dimensional description. We therefore believe that in high dimensions the outcome depends on which dimensions the map is computed in and on how the dimension is reduced, and we leave both to future work.

\section{Related Work}
\label{sec:related}

\subsection{Surrogate-Assisted Evolutionary Optimization}
When an evaluation takes minutes or hours, the number of affordable evaluations is a few hundred to a few thousand, and an EA usually relies on a model of the objective~\cite{jin2011surrogate,jin2019datadriven,li2022ecexpensive,zhou2024highdimexpensive}. Kriging (Gaussian-process) models with infill criteria such as expected improvement~\cite{jones1998ego,garnett2023bo} and radial basis function (RBF) models~\cite{gutmann2001rbf,regis2007srbf} are the most common surrogates. DYCORS~\cite{regis2013dycors} perturbs the best point found so far, each coordinate with a probability that decreases over the run, and selects among the perturbations by a weighted score of predicted value and distance to the evaluated points; it remains a strong reference for budgets of $50D$ to $1000$ evaluations (implementation in~\cite{eriksson2019pysot}). TuRBO~\cite{eriksson2019turbo} runs Bayesian optimization inside trust regions whose size follows a similar success/failure rule, and lq-CMA-ES~\cite{hansen2019lqcmaes} lets a global linear-quadratic model decide which offspring of CMA-ES~\cite{hansen2001cma} are evaluated. Surrogate-assisted EAs combine such models with population-based search~\cite{zhen2023twostage,wu2023ensemble,liu2024sadekt}. GPEME~\cite{liu2014gpeme} pre-screens offspring with a Gaussian process in a reduced space. For high-dimensional problems (tens to hundreds of variables), several methods confine the model or the search to low-dimensional subspaces: LineBO~\cite{kirschner2019linebo} optimizes the acquisition function on one-dimensional lines through the best point, BOIDS~\cite{ngo2025boids} selects among lines guided by the incumbents and combines them with subspace embeddings, and RPHSA~\cite{ren2021rphsa} averages RBF models trained on random low-dimensional projections of the samples. Our contour map is built in the full space; its $n$-dimensional level sets are measured along rays from a peak, a one-dimensional probing of the surrogate that serves to describe a peak, not to choose the next point. SA-COSO~\cite{sun2017sacoso} lets a surrogate-assisted PSO and a social-learning PSO cooperate. CAL-SAPSO~\cite{wang2017calsapso} uses a committee of surrogates for active learning, SHPSO~\cite{yu2018shpso} organizes the search hierarchically, the level-based learning swarm optimizer of~\cite{wei2021llso} replaces regression by classification, and the sampling agent of~\cite{zhen2023esa} chooses among sampling strategies by reinforcement learning. Global and local surrogates are combined in~\cite{zhou2007memetic}. For expensive multimodal problems, MaMPSO~\cite{ji2023msmtpso} treats three surrogates as the tasks of a multitasking niching PSO and confines trust-region local search to the best solution of each modality, and DSADE~\cite{ji2024multiplesolutions} searches density-clustered subregions with separate subpopulations. In all these methods the surrogate scores individual candidate points. The present work instead fits an interpolated surface to the evaluated solutions and uses it as a map whose topology (which peaks exist and how prominent each is) and whose contour geometry (how the region above a level shrinks) are read explicitly, and from which the location, height and shape of each peak are estimated.

\subsection{Topology of Landscapes and Basin Identification}
Topographical global optimization (TGO)~\cite{torn1994tgo} keeps the sampled points that are better than their $k$ nearest neighbors as starting points for local search. The simplicial homology global optimization (SHGO) method~\cite{endres2018shgo} identifies locally minimal vertices of a triangulation and uses homology groups to decide when the search can stop. Nearest-better clustering~\cite{preuss2012nbc} and many niching methods~\cite{li2017niching,liang2024ncide} identify basins from the relative positions of good points, and exploratory landscape analysis~\cite{mersmann2011ela,seiler2025deepela} summarizes the topography by features computed from samples. Hill-valley clustering~\cite{maree2021hillvalley} processes the samples in order of fitness and joins each one to the cluster of a nearby better sample unless an evaluated test point between them is worse than both. Two recent lines of work recover the peak structure explicitly. Persistence-based clustering~\cite{chazal2013tomato} climbs a neighborhood graph of sample points to the modes of a function and merges the modes whose persistence is below a threshold, and Meng et al.\ apply it to the population of an EA. CECA~\cite{meng2024ceca} uses it inside chaotic evolution and refines each cluster by Gaussian local search. SRS-GMO~\cite{meng2026srsgmo} and CLDE~\cite{meng2026mapping} build a neighborhood graph on the current samples, assign each sample to a peak through its best neighbor, merge two peaks when the valley between them is shallower than a threshold, and allocate the search effort among the basins by scores that include their depth, a persistence-style rule applied to the samples themselves. APDMMO~\cite{ma2025apdmmo} runs gradient descent on a neural surrogate from many random starts, clusters the end points and starts a local search from each cluster. Peak identification after the search~\cite{zhang2024peakid} separates the final solutions with hill-valley tests. Barrier trees~\cite{flamm2002barrier} and local optima networks~\cite{ochoa2008lon,mitchell2023lonbbob} describe how the basins of a landscape are connected. Statistical estimates of the optimum value from random samples rest on the volume law $V\propto(f^*-c)^{n/2}$ near a nondegenerate optimum~\cite{dehaan1981estimation,zhigljavsky1991theory}, which is also the leading term of our area law.

Contour lines have also been used inside EAs. In two dimensions, the contour method of~\cite{lin2008contour} triangulates the population, takes every individual that is better than all its neighbors in the triangulation as a local optimum, extends the triangle edges beyond it to one contour level slightly above its fitness by linear extrapolation, and evaluates the centroid of the points so obtained; in more dimensions the neighborhoods are the nearest neighbors of the best individuals. ANDE~\cite{wang2020ande} applies the same construction within the niches of a niching DE. Both extrapolate one contour level linearly around a local optimum and estimate neither height nor shape. This paper adds the decomposition of the evaluated region into peaks by a persistence-pruned tree, a memory of the explored peaks and, to describe each peak, the nested band and the area law. The landscape-aware DE of~\cite{lin2026lade} approximates the region of each peak found so far by an axis-aligned box spanned by the worse historical individuals around it, treats these boxes as taboo regions, and separates new peaks from known ones with hill-valley tests~\cite{maree2021hillvalley}. None of these methods estimates the height or the shape of a peak, and the contours of the first two consist of a single level around each locally best point. In computational topology the contour tree~\cite{carr2003contourtree} records how the components of the level sets of a function appear, merge and disappear, and persistence~\cite{edelsbrunner2002persistence,edelsbrunner2010book,dey2022ctda} measures the prominence of each extremum as the difference between its value and the level at which it merges with a higher one. In two dimensions we use the merge tree of the superlevel sets of an interpolated surface, the part of the contour tree that records how peaks merge, together with persistence to find peaks. Unlike the methods above, we also extrapolate each peak's nested contours to estimate its height, location and shape; in our experiments these estimates are accurate near the top of smooth peaks, and as start points for the search the simpler maximum of the interpolated surface is at least as good (Sections~\ref{sec:exp-est} and~\ref{sec:exp-abl}).

\subsection{Memetic Algorithms}
Memetic algorithms combine global evolutionary search with local refinement~\cite{ong2004metalamarckian,krasnogor2005memetic,neri2012memetic}. Their design questions are where, when and how intensively to apply local search. A contour step answers them from the map (Section~\ref{sec:caea}). Keeping a memory of the optima already found, so that later effort goes elsewhere, is also the purpose of the repelling subpopulations of~\cite{ahrari2022rscmsa}, which reject samples falling inside ellipsoidal taboo regions around the archived optima, of the taboo peak regions of~\cite{lin2026lade}, and of the optimal-peak detection of~\cite{jie2023opdde}, which archives the best individual of a niche and reinitializes the niche once this individual and the best fitness of the population have both stagnated and agree within a tolerance; in a contour step the memory is attached to the peaks of the contour tree. Nelder--Mead~\cite{nelder1965simplex}, with the adaptive coefficients of~\cite{gao2012adaptivenm} above three dimensions, serves as the derivative-free local search.

\section{Estimating Peaks From Contour Lines}
\label{sec:method}

\begin{figure*}[!t]
\centering
\includegraphics[width=0.76\textwidth]{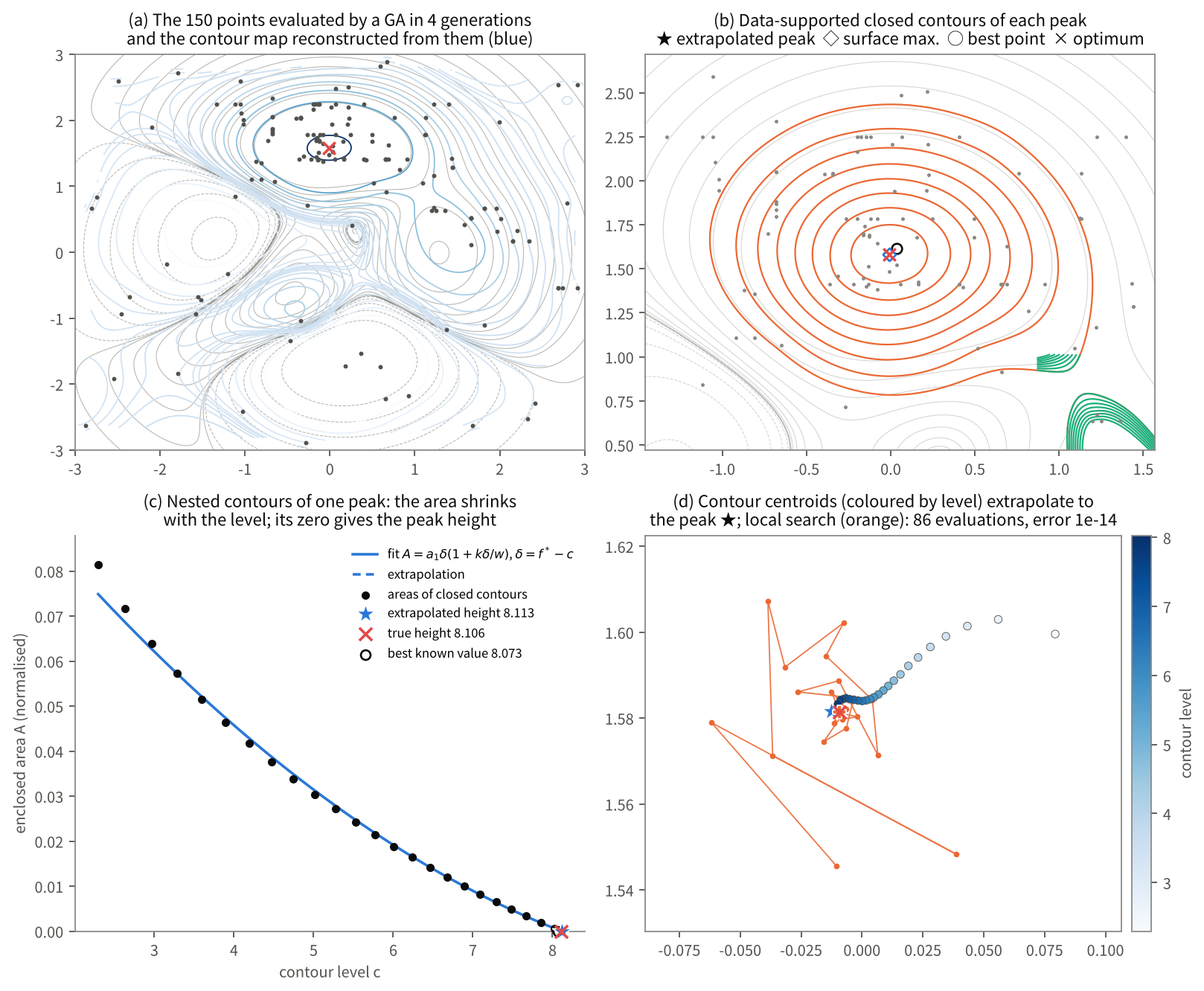}
\caption{One contour step on the Peaks function after four generations of a GA. (a) The 150 evaluated points and the contour map reconstructed from them (blue) over the true contours (grey). (b) Data-supported closed contours of the significant peaks. (c) Areas enclosed by the nested contours of the main peak versus the contour level, and the fitted area law; the zero of the fitted curve is the estimated peak height (8.113; true height 8.106; best evaluated value 8.073). (d) Centroids of the nested contours, colored by level, extrapolate to the estimated peak (star). A Nelder--Mead search started there (orange) reaches the global maximum to $10^{-14}$ in 86 evaluations.}
\label{fig:demo}
\end{figure*}

We consider the maximization of $f:\Omega\to\R$ over a box $\Omega\subset\R^d$; minimization problems are negated. All computations use normalized coordinates $\bu\in[0,1]^d$. The archive $\mathcal{A}=\{(\bu_i,y_i)\}_{i=1}^{N}$ contains every solution evaluated so far, for instance all individuals generated by an EA. Before a map is drawn, points closer than $10^{-3}$ to a better point are removed, and the spacing is enlarged until at most $N_{\max}=\min(100d+200,600)$ points remain; because the better point of every cluster is kept, the high ground stays densely sampled.

\subsection{Contour Map From Evaluated Solutions}
\label{sec:map}
\emph{Value compression.} Landscapes with steep walls make interpolants oscillate near their peaks. Since contour lines are invariant under strictly increasing transformations of $f$, we interpolate the compressed values
\begin{equation}
t_i=-\ln\!\Big(1+\frac{y_{\max}-y_i}{s}\Big),\qquad s=\operatorname{median}_i\,(y_{\max}-y_i),
\label{eq:compress}
\end{equation}
which preserve the order of the values, keep the high ground nearly linear and compress the deep valleys. If the median is zero, $s$ is the standard deviation of the values $y_i$, and $s=1$ if that is zero as well. All contour levels are mapped back to the original scale before any extrapolation.

\emph{Surface.} The surface $\hat s$ is the RBF interpolant
$\hat s(\bu)=\sum_{i} w_i\,\phi(\|\bu-\bu_i\|)+p(\bu)$ with a linear polynomial $p$, the thin-plate spline $\phi(r)=r^2\ln r$ for $d=2$ and the polyharmonic spline $\phi(r)=r^3$ for $d>2$~\cite{duchon1977tps,wendland2004scattered}. The thin-plate spline is the surface of minimum bending energy through the data, the classical choice for contouring scattered spot heights.

\emph{Mapped area.} Contours are trusted only inside a mapped area $\mathcal{M}$. For $d>2$, $\mathcal{M}=\bigcup_i B(\bu_i,\rho_i)$, where $\rho_i$ is the distance from $\bu_i$ to its $(d+1)$-th nearest neighbor; this excludes the large empty regions that the convex hull of a few hundred points contains in three or more dimensions (with a few hundred points in 30 dimensions the balls exclude little, and the thinness of the band is the effective safeguard, Section~\ref{sec:exp-abl}). For $d=2$, where the archive of an EA quickly spreads over the whole box and the convex hull contains few empty regions, $\mathcal{M}$ is the convex hull of the evaluated points together with disks of radius 2.5 grid cells around them, a cheap test that keeps every evaluated point inside. A contour line is \emph{closed} if the region it encloses does not touch the boundary of $\mathcal{M}$ or of the unit box.

\subsection{Merge Tree and Peaks ($d=2$)}
\label{sec:tree}
The surface is evaluated on a $G\times G$ grid ($G=161$). Let $c_1<\dots<c_{n_\ell}$ ($n_\ell=120$) be levels evenly spaced between the 30\% quantile and the maximum of $\hat s$ over $\mathcal{M}$. For every level we label the 8-connected components of the superlevel set $\{\hat s\ge c_k\}\cap\mathcal{M}$. Processing the levels from the top down yields the merge (join) tree of the superlevel sets, the part of the contour tree~\cite{carr2003contourtree} that records how peaks merge; related barrier trees describe the basins of discrete landscapes~\cite{flamm2002barrier}. A component that appears at some level is a new peak; when the components of several peaks merge, the merged component keeps the identity of the highest peak, and the others die at the merge level (the elder rule). The persistence of a peak is its height minus the level at which it dies~\cite{edelsbrunner2002persistence}. A peak is \emph{significant} if its persistence exceeds 1\% of the level range (at least two levels); this removes interpolation ripples. Significance is a property of the interpolated surface: a significant peak is a candidate supported by the data, which the verification and local search of Section~\ref{sec:algorithms} confirm or reject. We therefore distinguish a \emph{peak of the surface} (a candidate), a \emph{verified peak} (the end point of a converged local search on $f$) and the global optimum. Finitely many evaluations cannot exclude a higher peak in a region without samples, so the map yields candidates, not certificates, and no upper bound on the heights of the peaks of $f$. The chain of a significant peak is the sequence of its components from its top down to its highest merge with another significant peak; the lowest component of the chain is the peak's \emph{basin}, and the chain prefix that stays closed in $\mathcal{M}$ contains its \emph{trusted contours}.

\subsection{Data-Supported Band}
\label{sec:band}
Contours near the top of a peak enclose few or no evaluated points; their shape is dictated by the interpolant alone. We therefore extrapolate only from a band of contours supported by data. Let $y_{(1)}\ge y_{(2)}\ge\cdots$ be the values of the evaluated points in the basin. The band is $[c_{\mathrm{lo}},c_{\mathrm{hi}}]$, where $c_{\mathrm{lo}}$ is the lowest trusted contour and $c_{\mathrm{hi}}=t(y_{(m_b)})$ with $m_b=2$, so every contour of the band encloses at least $m_b$ evaluated points. On a $121\times121$ grid zoomed onto the basin we take $n_b=24$ levels in the band; for each level we extract the component that contains the peak and, if it is closed in $\mathcal{M}$, measure its area $A(c)$, centroid $\bg(c)$ and second-moment matrix $\bS(c)$. Cells crossed by the contour are weighted by their estimated covered fraction: with $\eta=\tfrac12+(\hat s-c)/\|\nabla\hat s\|$ in grid units, cells inside the contour receive the weight $\min\{1,\max\{\tfrac12,\eta\}\}$ and their outer neighbors $\min\{\tfrac12,\max\{0,\eta\}\}$, which makes the measurements smooth functions of $c$. At least five closed contours are required; otherwise the peak falls back to the surface maximum (Section~\ref{sec:fallback}).

\subsection{The Area Law and Contour Extrapolation}
\label{sec:arealaw}
The extrapolation rests on two results; proofs are given in Section~\SuppRef{s:proofs} of the supplementary material. They are stated for a general dimension $n$; in the algorithms $n=d$.

\begin{proposition}[Quadratic peak]
\label{prop:quadratic}
Let $f(\bx)=f^*-\tfrac12(\bx-\bx^*)^\top\bH(\bx-\bx^*)$ with $\bH\succ 0$ in $\R^n$, and let $\delta=f^*-c>0$. The superlevel set $\{f\ge c\}$ is an ellipsoid with volume $V(c)=\omega_n(2\delta)^{n/2}(\det\bH)^{-1/2}$, centroid $\bx^*$ and second-moment matrix $\bS(c)=\tfrac{2\delta}{n+2}\bH^{-1}$, where $\omega_n=\pi^{n/2}/\Gamma(\tfrac n2+1)$. In particular, $V(c)^{2/n}$ is an affine function of $c$ that vanishes at $c=f^*$; for $n=2$, $A(c)=2\pi\delta/\sqrt{\det\bH}$.
\end{proposition}

\begin{proposition}[Smooth peak]
\label{prop:smooth}
Let $f$ be four times continuously differentiable near a nondegenerate local maximum $\bx^*$ in the interior of the domain, with Hessian $-\bH$, $\bH\succ0$, and let $V(c)$ and $\bg(c)$ be the volume and centroid of the component of $\{f\ge c\}$ that contains $\bx^*$. As $\delta=f^*-c\to0^+$,
\begin{align}
V(c)^{2/n}&=a_1\,\delta\,\big(1+k\delta+o(\delta)\big),\label{eq:areaexp}\\
\bg(c)&=\bx^*+\boldsymbol\beta\,\delta+o(\delta),\label{eq:centexp}
\end{align}
with $a_1=2\,\omega_n^{2/n}(\det\bH)^{-1/n}$ and constants $k$ and $\boldsymbol\beta$: $\boldsymbol\beta$ depends only on $\bH$ and the third derivatives of $f$ at $\bx^*$ and vanishes when they vanish, and $k$ depends on $\bH$ and the third and fourth derivatives (integral expressions for $k$ and $\boldsymbol\beta$ over the unit sphere are given in Section~\SuppRef{s:proofs} of the supplementary material).
\end{proposition}

In the proof, the leading correction to the radius of the level set, in polar coordinates around $\bx^*$, is an odd function of the direction. It therefore cancels to first order in the volume ($1+k\delta$ collects the even fourth-order terms and the square of the odd third-order term). In the centroid the symmetric leading term cancels instead, and this odd correction is the first term that remains, so the centroid moves linearly with $\delta$. Note that the thin-plate spline is not twice differentiable at its centers, so Proposition~\ref{prop:smooth} describes the contours of $f$ that the interpolant approximates rather than those of the interpolant itself. Proposition~\ref{prop:smooth} holds for levels close enough to $f^*$ that the component around $\bx^*$ is isolated; it does not cover maxima on the boundary of the domain, degenerate maxima or saddle points (closed forms of $k$ and $\boldsymbol\beta$ are given in supplementary Section~\SuppRef{s:proofs}). More generally, the propositions describe ideal level sets; the estimator works on an interpolant fitted to sparse samples, discretized contours or rays and a least-squares fit, so its accuracy is an empirical matter (Table~\ref{tab:oracle}, supplementary Tables~\SuppRef{tab:oraclend} and~\SuppRef{tab:oneshot}).

\emph{Height.} Guided by \eqref{eq:areaexp}, we fit
\begin{equation}
A(c)=a_1\,\delta\,\big(1+\tilde k\,\delta/w\big),\qquad \delta=f^*-c,
\label{eq:taylor2}
\end{equation}
to the measured areas (or $V^{2/n}$ for $n>2$) by least squares on $\ln A$, where $w$ is the width of the band in the original units and $\tilde k$ plays the role of $kw$ in \eqref{eq:areaexp}. The fit is constrained by $f^*\ge y_{(1)}$, since a peak cannot be lower than a point evaluated in its basin, by $f^*\le c_{\mathrm{top}}+2w$, and by $\tilde k\ge\tilde k_{\min}$ with $\tilde k_{\min}=-0.95\,w/(c_{\mathrm{top}}+2w-c_{\mathrm{bot}})$, which keeps the model positive on the band ($c_{\mathrm{top}}$ and $c_{\mathrm{bot}}$ denote the highest and lowest measured levels), with $a_1=e^{\ln a_1}>0$. They do not force $A$ to decrease over the whole band; \nMonoBad\ of the \nMonoFits\ fits recorded in supplementary Section~\SuppRef{s:impl} (\nMonoBadPct\%) are non-monotone. If $y_{(1)}>c_{\mathrm{top}}+2w$ the bounds conflict and $\hat f^*=y_{(1)}$. Residuals are taken in $\ln A$ so that small inner contours count as much as large outer ones. The cap $c_{\mathrm{top}}+2w$ is a heuristic limit of the extrapolation, not an upper bound on the height of the peak. The zero of the fitted model, $\hat f^*$, is the estimated peak height.

\emph{Location.} Guided by \eqref{eq:centexp}, the centroids are regressed linearly on $\delta=\hat f^*-c$ with weights proportional to $1/\delta$, which emphasizes the inner contours; the intercept is the extrapolated peak $\hat\bu^*$. If $\hat\bu^*$ falls outside the innermost measured contour, it is moved along the segment towards the surface maximum until it lies inside.

\emph{Shape.} The second-moment matrix of the innermost measured contour, $\hat\bS=\bS(c_{\mathrm{top}})$, describes the orientation and extent of the peak. By Proposition~\ref{prop:quadratic} it is proportional to $\bH^{-1}$ for a quadratic peak, so its eigenvectors are the principal axes of the peak.

\subsection{Fallbacks and Ranking}
\label{sec:fallback}
When a significant peak has no trusted closed contour, no supported band or too few contours, its estimate falls back to the maximum of $\hat s$ in its basin, refined by three nested $21\times21$ grid searches, with height $\max\{\hat s_{\max},y_{(1)}\}$ and a diagonal shape matrix sized from the basin. A peak whose basin contains no evaluated point is discarded. Each peak receives the score $\hat f^*$ when the extrapolation succeeds and $y_{(1)}$ otherwise, and peaks are ranked by their scores. Since the constrained height $\hat f^*$ is never below $y_{(1)}$, the ranking favors peaks whose contours support a height above their best evaluated point. Fig.~\ref{fig:demo} illustrates the procedure.

\subsection{Level Sets in $n$ Dimensions}
\label{sec:nd}
Grids are infeasible for $d>2$. A surveyor would describe a closed contour around a hilltop by its distance from the top in every direction, and we follow this description.

\emph{Peaks of the surface.} For $d>2$ we do not build a merge tree. The peaks are located by hill climbing on the surface: starting from the $n_s=\min\{40,\max\{2d,\lfloor0.1N\rfloor\}\}$ best evaluated points (60 in thirty dimensions while the map has fewer than 600 points), $\hat s$ is climbed by gradient ascent with adaptive steps (multiplied by 1.5 after a success and by 0.4 after a failure; 80 iterations), end points closer than $0.01\sqrt d$ are merged, and the (at most six) peaks with the highest surface values are kept. Instead of persistence, the separation between peaks is enforced by the base level of each peak's region described below.

\emph{Rays.} From a peak $\bu_s$ we cast $n_\theta=\max\{128,32d\}$ quasi-random unit directions $\boldsymbol\theta_k$ in antithetic pairs ($n_\theta=256$ in thirty dimensions, supplementary Table~\SuppRef{tab:params}) and sample $\hat s(\bu_s+r\boldsymbol\theta_k)$ at geometrically spaced radii. A ray stays valid while it is inside the unit box and inside $\mathcal{M}$. For a level $c$, the first-exit radius $R_k(c)$ is the first radius at which the profile drops below $c$ (linearly interpolated); it is undefined if the ray becomes invalid first. For the region that is star-shaped around $\bu_s$,
\begin{equation}
\begin{aligned}
V(c)&=\frac{|S^{d-1}|}{n_\theta}\sum_{k}\frac{R_k^d}{d},\\
\bg(c)&=\bu_s+\frac{\sum_k\boldsymbol\theta_kR_k^{d+1}/(d+1)}{\sum_kR_k^d/d},\\
\bS(c)&=\frac{\sum_k\boldsymbol\theta_k\boldsymbol\theta_k^\top R_k^{d+2}/(d+2)}{\sum_kR_k^d/d}-\mathbf m\mathbf m^\top,\quad \mathbf m=\bg-\bu_s .
\end{aligned}
\label{eq:rays}
\end{equation}
The lowest level at which every ray is closed (the escape level) and the highest of the minima of $\hat s$ along the straight segments to the other peaks give $c_{\mathrm{lo}}$ (the larger of the two, raised by a small margin so that the exits are strict). The rays measure the part of the superlevel set that is visible from $\bu_s$ along straight lines, a star-shaped subset of its connected component, and a segment minimum is only a lower bound on the saddle level between two peaks; hill climbing with segment minima is therefore a heuristic substitute for the persistence-pruned merge tree, not its equivalent in $d$ dimensions. The basin consists of the evaluated points whose segment from $\bu_s$ stays above $c_{\mathrm{lo}}$, and $c_{\mathrm{hi}}$ is defined as in two dimensions. Over $n_b=20$ levels of the band, \eqref{eq:taylor2} is fitted to $V^{2/d}$ and the centroids are extrapolated as before. With finitely many rays, \eqref{eq:rays} is a numerical quadrature, which misses the long axes of an ill-conditioned peak: for a quadratic peak in 30 dimensions with condition number $10^6$, the median estimated volume is \nRayVolTS\ of the exact one with $n_\theta=256$ (supplementary Section~\SuppRef{s:raystress}). On a quadratic peak this error is a constant factor in $V(c)$, so it biases $a_1$ and the shape matrix but not the estimated height and location. The shape matrix is therefore used only to orient sampling, with clipped eigenvalues (Section~\ref{sec:casaea}).

\subsection{Computational Cost}
\label{sec:cost}
Fitting the RBF costs $O(N^3)$ with $N\le N_{\max}$. In two dimensions, drawing the map costs $O(G^2N)$ operations, the merge tree $O(n_\ell G^2)$ and the band $O(n_bG_z^2+G_z^2N)$ with $G_z=121$. In $d$ dimensions, the peak search costs $O(n_sdN)$ operations per iteration and the rays $O(n_\theta n_rNd)$ operations per peak for $n_r=48$ to $64$ radii, for at most six peaks. The cost is independent of the cost of evaluating $f$ and is negligible when an evaluation takes seconds or more; measured times are reported in Section~\ref{sec:exp-cost}.

\section{Contour-Assisted Evolutionary Algorithms}
\label{sec:algorithms}

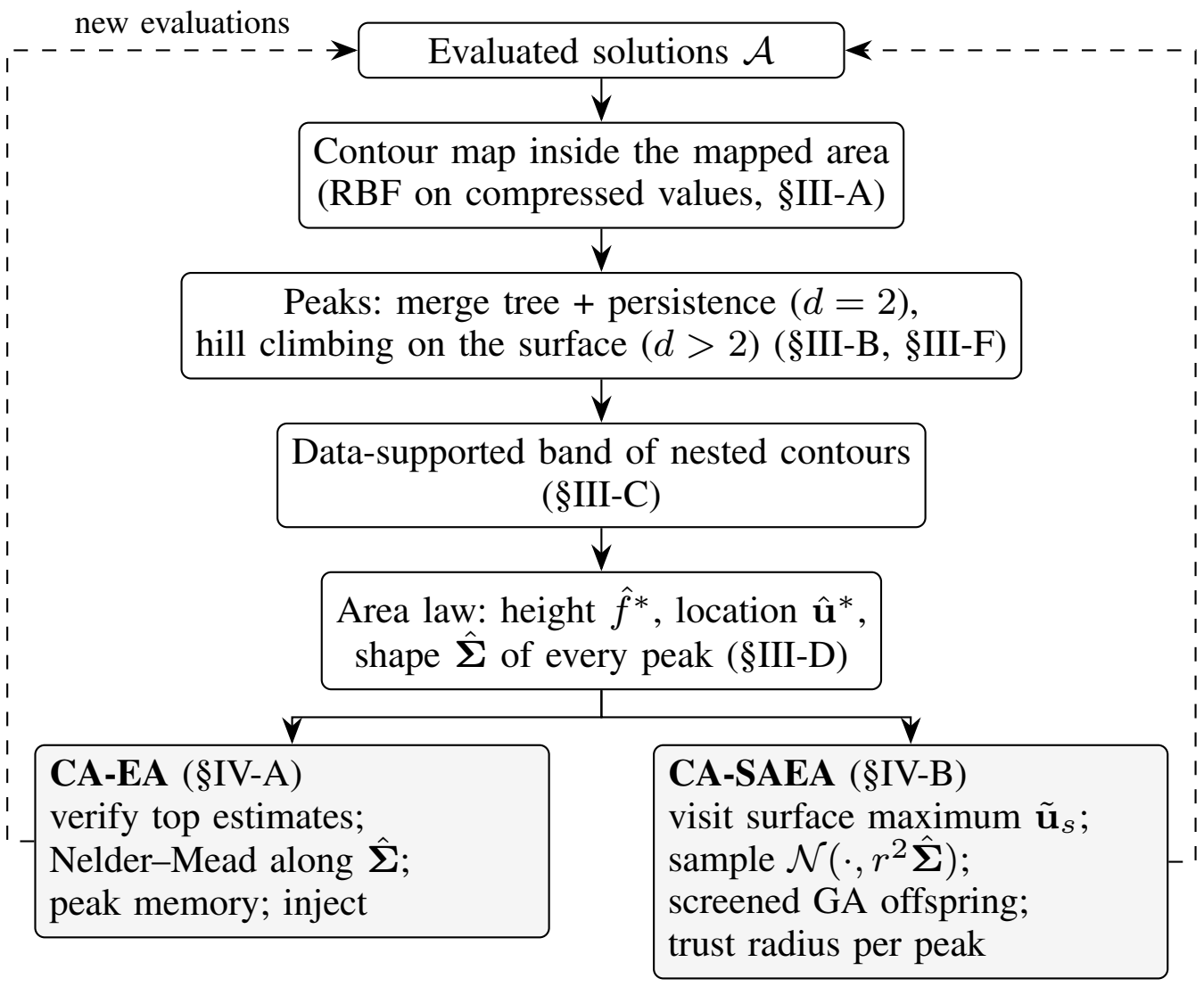
\begin{figure}[!t]
\centering
\begin{tikzpicture}[font=\footnotesize, node distance=3.2mm,
  box/.style={draw, rounded corners=2pt, align=center, inner sep=3pt, minimum width=36mm},
  alg/.style={draw, rounded corners=2pt, align=left, inner sep=3pt, text width=36mm, fill=black!4},
  arr/.style={-{Stealth[length=2mm]}}]
\node[box] (a) {Evaluated solutions $\mathcal{A}$};
\node[box, below=of a] (b) {Contour map inside the mapped area\\(RBF on compressed values, \S\ref{sec:map})};
\node[box, below=of b] (c) {Peaks: merge tree + persistence ($d=2$),\\hill climbing on the surface ($d>2$) (\S\ref{sec:tree}, \S\ref{sec:nd})};
\node[box, below=of c] (d) {Data-supported band of nested contours\\(\S\ref{sec:band})};
\node[box, below=of d] (e) {Area law: height $\hat f^*$, location $\hat\bu^*$,\\shape $\hat\bS$ of every peak (\S\ref{sec:arealaw})};
\node[alg, below left=4mm and -17mm of e] (f) {\textbf{CA-EA} (\S\ref{sec:caea})\\verify top estimates;\\Nelder--Mead along $\hat\bS$;\\peak memory; inject};
\node[alg, below right=4mm and -17mm of e] (g) {\textbf{CA-SAEA} (\S\ref{sec:casaea})\\visit surface maximum $\tilde\bu_s$;\\sample $\mathcal N(\cdot,r^2\hat\bS)$;\\screened GA offspring;\\trust radius per peak};
\draw[arr] (a)--(b); \draw[arr] (b)--(c); \draw[arr] (c)--(d); \draw[arr] (d)--(e);
\draw[arr] (e.south)--++(0,-2mm)-|(f.north); \draw[arr] (e.south)--++(0,-2mm)-|(g.north);
\draw[arr, dashed] (f.west)--++(-2mm,0)|-(a.west) node[pos=0.75, above, font=\scriptsize]{new evaluations};
\draw[arr, dashed] (g.east)--++(2mm,0)|-(a.east);
\end{tikzpicture}
\caption{Contour-assisted evolutionary computation: the solutions evaluated so far are turned into a contour map, whose peaks are estimated and then exploited by CA-EA (mid-run acceleration of an EA) or CA-SAEA (expensive optimization).}
\label{fig:framework}
\end{figure}

Fig.~\ref{fig:framework} summarizes how the peak estimates are used. CA-EA keeps an ordinary EA in charge and occasionally inserts a contour step; CA-SAEA is designed for budgets of a few hundred evaluations, in which every evaluation must be chosen with care.

\subsection{CA-EA: Contour Steps in the Middle of an Evolutionary Run}
\label{sec:caea}
CA-EA (Algorithm~\ref{alg:caea}) runs a GA, DE or PSO unchanged and, every $T$ generations after a warm-up of $W$ generations, performs a contour step (Algorithm~\ref{alg:step}) on the archive of all evaluated solutions. The step estimates the peaks and discards those already explored. It then evaluates the estimated peak points of the $k_v$ best-ranked peaks, which costs at most $k_v$ evaluations, and selects, among these verified peaks, the peak $j$ with the largest $\max\{\mathrm{score}_j,f(\hat\bu^*_j)\}$, where $\mathrm{score}_j$ is the ranking score of Section~\ref{sec:fallback}. A Nelder--Mead search starts from the selected estimate, or from the best point of the peak's basin if that point is better; this rule, rather than the selection, protects the search against a poor estimate. The initial simplex is aligned with the eigenvectors $\mathbf v_i$ of the shape matrix $\hat\bS$, with edge lengths $\sqrt{\lambda_i}$ clipped to $[10^{-3},0.25]$, so the first simplex already has the orientation and scale of the peak. The search is limited to $m_{LS}$ evaluations; a search that stops at this limit is marked unconverged and may be resumed by a later step. All evaluations pass through one counter, so a step stops as soon as the budget $B$ is used.

\emph{Peak memory.} A peak is \emph{explored} if its estimated peak lies within $r_e=0.02\sqrt{d/2}$ of the start or end point of an earlier converged search, or if its region contains the end point of an earlier search whose value is at least the peak's score. Explored peaks are skipped, so successive steps climb different peaks. The results of the step replace the worst individuals of the population if they are better and not already present; in PSO the replaced particles also receive the new position and zero velocity.

\begin{algorithm}[!t]
\caption{CA-EA}
\label{alg:caea}
\begin{algorithmic}[1]
\Require EA (GA, DE or PSO), budget $B$, warm-up $W$, period $T$
\State initialize the population; $g\gets0$
\While{fewer than $B$ evaluations used}
  \State run one generation of the EA; $g\gets g+1$
  \If{$g\ge W$ \textbf{and} $(g-W)\bmod T=0$}
    \State $\mathcal{R}\gets$ \Call{ContourStep}{$\mathcal{A}$}
    \State replace the worst individuals by the better points of $\mathcal{R}$
  \EndIf
\EndWhile
\State \Return the best evaluated solution
\end{algorithmic}
\end{algorithm}

\begin{algorithm}[!t]
\caption{ContourStep (used by CA-EA)}
\label{alg:step}
\begin{algorithmic}[1]
\Require archive $\mathcal{A}$, memory $\mathcal{E}$ of earlier searches, $k_v$, $m_{LS}$
\State thin $\mathcal{A}$; estimate peaks $\{(\hat\bu^*_j,\hat f^*_j,\hat\bS_j)\}$ (Section~\ref{sec:method})
\State remove the peaks that are explored according to $\mathcal{E}$
\State evaluate $f(\hat\bu^*_j)$ for the $k_v$ best-ranked peaks
\State $j^\star\gets\arg\max_j \max\{\mathrm{score}_j, f(\hat\bu^*_j)\}$ over the verified peaks
\State $\bu_0\gets\hat\bu^*_{j^\star}$, or the best point of the basin if it is better
\State build the simplex $\{\bu_0\}\cup\{\bu_0+\sqrt{\lambda_i}\,\mathbf v_i\}$ from $\hat\bS_{j^\star}$
\State run Nelder--Mead for at most $m_{LS}$ evaluations; add its start, end and convergence flag to $\mathcal{E}$
\State \Return the end point and the other evaluated estimates
\end{algorithmic}
\end{algorithm}

\subsection{CA-SAEA: Contour-Assisted Expensive Optimization}
\label{sec:casaea}
With a budget of $50D$ evaluations, two features of CA-EA become wasteful: the EA spends whole generations on offspring that no model has screened, and Nelder--Mead needs many evaluations in ten or more dimensions. CA-SAEA (Algorithm~\ref{alg:casaea}) therefore selects every evaluation with an RBF surrogate of the compressed values~\eqref{eq:compress}, refitted after each evaluation, and alternates two kinds of steps: in every cycle of $L+1$ steps, $L$ steps are local and one is global, and a step is also global whenever the trackers of all peaks are explored. In the runs of Sections~\ref{sec:exp-mq} and~\ref{sec:exp-dim} with $L=4$, \nvGlobTen\% of the steps of CA-SAEA were global in ten dimensions (between \nvGlobMinTen\% and \nvGlobMaxTen\% depending on the function) and \nvGlobThirty\% in thirty.

\emph{Local steps (exploitation).} The contour map is redrawn every $k_c$ evaluations. Each peak carries a tracker that holds the best point $\mathbf b_s$ found in its basin, a trust radius $r_s$ and counters of consecutive successes and failures. When the map is redrawn, a new peak inherits the tracker of an old one if its region contains the old tracker's best point or if its estimate lies within $r_e$ of it. Otherwise it receives a new tracker with $r_s=r_0$ and $\mathbf b_s$ set to the best point of its basin. Peaks are ranked by the maximum of the interpolated surface in their basin (or by the best evaluated value in the basin if that is higher), and a local step visits a peak whose tracker is not explored: the highest-ranked one, except at the beginning of the run, when several peaks are kept (see below). The first visit to a peak after each redrawing evaluates the location $\tilde\bu_s$ of its surface maximum itself. Later visits draw $n_c$ candidates around $\mathbf b_s$. Half of them are Gaussian with covariance $r_s^2\hat\bS_s$, whose eigenvalues are clipped to $[10^{-6},0.0625]$, so they follow the shape of the peak's innermost contour. The other half are sparse coordinate perturbations with the same per-axis scales, in which each coordinate is perturbed with the decreasing probability used by DYCORS~\cite{regis2013dycors}. The candidate with the best merit (see below) is evaluated. After three consecutive successes (a success is any improvement of $\mathbf b_s$) the radius is doubled, up to $r_{\max}$; after $\max\{d,4\}$ consecutive failures it is halved. A peak whose radius falls below $r_{\min}$ is marked explored. This is the step-size rule of DYCORS~\cite{regis2013dycors}, which trust-region Bayesian optimization also uses~\cite{eriksson2019turbo}; what is new is that every peak has its own radius and memory. The success rule uses only comparisons of function values, and the value compression~\eqref{eq:compress} only their differences, so CA-SAEA is invariant to adding a constant to $f$.

\emph{Keeping several peaks first.} The first maps are drawn from few solutions, and the peak that they rank highest is often not the highest peak of the landscape. A search that follows only this peak seldom leaves it later: the evaluations spent on it raise its rank further, and what the early solutions tell about the other peaks is no longer used. CA-SAEA therefore begins with up to $m$ candidates, the $m$ highest-ranked peaks whose trackers are not explored, and its local steps visit them in turn. From the second map on, every new map reduces the number of candidates by one, so that after $m-1$ redrawings a single peak is followed as described above. While more than one candidate is kept, the map is drawn from more starting points: the hill climbing of Section~\ref{sec:nd} starts from the $n_s$ best points and also from the locally best points that are not among them, where a point is locally best if none of its $k_\ell$ nearest evaluated points is better (at most $\max\{n_s,40\}$ locally best points are considered, the best first). Peaks none of whose points is among the $n_s$ best then appear on the map as well. We use $m=6$ and $k_\ell=3$ (Section~\ref{sec:exp-setup}).

\emph{Global steps (exploration).} A GA (binary tournament, simulated binary crossover~\cite{deb1995sbx} and polynomial mutation) breeds $n_o$ offspring from the $N_P$ best evaluated points, and the candidate with the best merit is evaluated.

\emph{Merit.} As in DYCORS, a candidate $\bu$ is scored by
$m(\bu)=\nu\,\tilde s(\bu)+(1-\nu)\,(1-\tilde\Delta(\bu))$,
where $\tilde s$ is the predicted value rescaled to $[0,1]$ over the candidates ($0$ for the best prediction; the candidate with the smallest $m$ is evaluated) and $\tilde\Delta$ is the distance to the nearest evaluated point rescaled to $[0,1]$. The weight $\nu$ cycles through $0.3$, $0.5$, $0.8$ and $0.95$, which alternates between exploration and exploitation, and candidates closer than $10^{-3}$ to an evaluated point are excluded. The surrogate is fitted on at most $M$ points: the points nearest to $\mathbf b_s$ for local steps and the best points for global steps.

\begin{algorithm}[!t]
\caption{CA-SAEA}
\label{alg:casaea}
\begin{algorithmic}[1]
\Require budget $B$, $n_0$, $L$, $k_c$, $m$, $n_c$, $n_o$, $N_P$, $M$, $r_0$, $r_{\min}$, $r_{\max}$
\State evaluate a Latin hypercube design~\cite{mckay1979lhs} of $n_0$ points; $t\gets0$; $m'\gets m$
\While{fewer than $B$ evaluations used}
  \If{$k_c$ evaluations since the last map} \State redraw the map; estimate peaks; match trackers; $m'\gets\max\{m'-1,1\}$ unless this is the first map \EndIf
  \State $s\gets$ the next in turn of the $m'$ highest-ranked peaks with unexplored trackers
  \If{$s$ exists \textbf{and} $t\bmod(L+1)\ne L$} \Comment{local step}
     \If{$\tilde\bu_s$ not yet evaluated for this map} \State $\bu\gets\tilde\bu_s$
     \Else \State fit $\hat s$ on the $M$ points nearest to $\mathbf b_s$; draw $n_c$ candidates around $\mathbf b_s$ with shape $r_s^2\hat\bS_s$; $\bu\gets$ best merit \EndIf
     \State evaluate $f(\bu)$; update $\mathbf b_s$, the counters and $r_s$; mark $s$ explored if $r_s<r_{\min}$
  \Else \Comment{global step}
     \State fit $\hat s$ on the $M$ best points; breed $n_o$ GA offspring of the $N_P$ best points; $\bu\gets$ best merit; evaluate $f(\bu)$
  \EndIf
  \State $t\gets t+1$
\EndWhile
\State \Return the best evaluated solution
\end{algorithmic}
\end{algorithm}

\emph{Variants.} CA-SAEA (one peak) is CA-SAEA with $m=1$: it follows the highest-ranked peak from the first map on and draws every map from the $n_s$ best points only. This was the algorithm before the candidates were introduced (Section~\ref{sec:exp-setup}), and the comparison with it shows what keeping several peaks contributes. Two further variants isolate the contribution of the contour map. CA-SAEA (extrap.) ranks the peaks by their extrapolated heights and enters each peak at its extrapolated location $\hat\bu^*$ instead of at the surface maximum; this is how our initial design entered the peaks. CA-SAEA (best) draws no contour map at all: its local steps sample around the best point found so far with the isotropic shape $\sigma=0.1$ (in normalized coordinates) and a single tracker, which makes it close to DYCORS combined with surrogate-screened GA offspring. Because its tracker restarts around the best point with radius $r_0$ when its radius falls below $r_{\min}$, its global steps are exactly one in $L+1$, whereas CA-SAEA and CA-SAEA (extrap.) take additional global steps when all their peaks are explored; the comparison therefore also varies the share of global steps.

\section{Experimental Study}
\label{sec:experiments}

The experiments address five questions. How accurate are the peak estimates (Section~\ref{sec:exp-est})? How much does a contour step accelerate an EA (Section~\ref{sec:exp-caea})? How does CA-SAEA compare with established and recent surrogate-assisted algorithms on multi-peak landscapes when only $50d$ evaluations are available (Section~\ref{sec:exp-mq})? How does the comparison change with the dimension (Section~\ref{sec:exp-dim})? Which components produce the gains, how sensitive are they to their parameters, and what do they cost (Sections~\ref{sec:exp-abl}--\ref{sec:exp-cost})?

\subsection{Setup}
\label{sec:exp-setup}
\emph{Test problems.} The two-dimensional suite contains nine functions in maximization form: Peaks (three maxima), Himmelblau (four equal maxima), six-hump camel, Branin, two random landscapes of anisotropic Gaussian peaks in the style of~\cite{gallagher2006gaussian} with 10 and 21 peaks (GaussPeaks and GaussPeaks21; global height 10 and other heights uniform in $[1.1,9.1]$, a new random landscape in every run), Rosenbrock, Ackley and Rastrigin. The five- and ten-dimensional suites contain GaussPeaks (21 peaks), Rosenbrock, Ackley and Rastrigin. For expensive optimization we use multi-peak landscapes of our own design and classic functions, which are described in Section~\ref{sec:exp-mq}, in five and ten dimensions and, for a subset, in twenty and thirty.

\emph{Protocol.} The budgets are 2000 evaluations (2D, 30 runs), $1000d$ evaluations (5D and 10D, 20 runs) and, for expensive optimization, $50d$ evaluations (20 runs). On the first three suites we report the expected running time (ERT)~\cite{hansen2021coco} to reach a relative error $(f^*-f)/(f^*-f_{\mathrm{med}})\le10^{-6}$, where $f_{\mathrm{med}}$ is the median of $f$ under uniform sampling, together with the success rate (SR). The ERT is the number of evaluations spent by all runs, up to the target in a successful run and the whole budget otherwise, divided by the number of successful runs (infinite when no run succeeds). For expensive optimization we report the error $f^*-f(\bx)$ after the budget. Differences are tested with the Wilcoxon signed-rank test for paired runs (algorithms that share the seed and the initial design, hence the initial population) and with the Wilcoxon rank-sum test otherwise, at $\alpha=0.05$, with Holm's correction~\cite{holm1979} where multiple comparisons are pooled~\cite{derrac2011tests}.

\emph{Algorithms.} The EAs are a real-coded GA (binary tournament, simulated binary crossover with $\eta_c=15$ and $p_c=0.9$, polynomial mutation with $\eta_m=20$ and $p_m=1/d$, $(\mu+\lambda)$ survival), DE/rand/1/bin ($F=0.5$, $CR=0.9$)~\cite{storn1997de} and global-best PSO with constriction coefficients~\cite{kennedy1995pso,clerc2002constriction}. Four controls insert a Nelder--Mead search at the same moments as CA-GA and differ from it in one or more components. GA+LS(surface) uses the same merge tree, peak memory and shape-aware simplex as CA-GA but starts from the maximum of the interpolated surface in the selected peak's region (no extrapolation). CA-GA(iso) is CA-GA with an isotropic initial simplex whose edge length is the geometric mean of the shape-aware edge lengths (no shape). GA+LS(topo) starts every search from the best evaluated point that is better than its $2d+2$ nearest neighbors and not yet explored, as in topographical global optimization~\cite{torn1994tgo}, with an isotropic simplex (no contour map, but restarts at distinct locally best points of the archive). GA+LS(best), the classical memetic control, starts from the best point found so far with an isotropic simplex of edge length $\min\{0.1,\max\{10^{-3},d_1/2\}\}$, where $d_1$ is the distance from the best point to its nearest evaluated neighbor; because the best point changes only when the EA improves it, this control performs far fewer local searches than CA-GA (Section~\ref{sec:exp-caea}). DE+LS(best) and PSO+LS(best) are defined in the same way. References are IPOP-CMA-ES~\cite{auger2005ipop} (the reference implementation of its authors, $\sigma_0=0.3$ in normalized coordinates), multistart Nelder--Mead, DYCORS~\cite{regis2013dycors} (our re-implementation with the default settings of~\cite{eriksson2019pysot}), and the surrogate-assisted evolutionary algorithms ESA~\cite{zhen2023esa} and TS-DDEO~\cite{zhen2023twostage}, re-implemented from the programs released by their authors and checked against the published results (supplementary Section~\SuppRef{sec:supp-recent}). ESA and TS-DDEO are run with the authors' default parameters and, in five and ten dimensions, also with 50 instead of the default 100 initial samples; for each of the two algorithms and each of these two dimensions we report the setting with the lower mean rank of the median error over the functions of that dimension, a rule fixed before the runs (50 samples for ESA in five dimensions, the default otherwise). In twenty and thirty dimensions the default is used. Supplementary Table~\SuppRef{tab:params} lists all parameters.

\emph{Configuration selection.} The parameters of the peak estimator (supplementary Table~\SuppRef{tab:params}) were fixed during the design of the method. The warm-up, period and number of verified peaks of CA-EA were chosen in development runs on five functions of the 2D suite (Peaks, Himmelblau, GaussPeaks, Ackley and Rastrigin) among $W\in\{3,5,10\}$, $T\in\{5,10\}$ and $k_v\in\{1,2,3\}$; these runs used run indices and random landscapes disjoint from those of the reported runs, but for the deterministic functions this amounts to tuning on test functions with different seeds, so the 2D results of CA-EA may be slightly optimistic (all variants with mid-run local search share $W$ and $T$). The settings of CA-SAEA for the budget of $50d$ evaluations (initial design and local-to-global ratio) were chosen on five development functions of our own (a rotated ellipsoid and shifted Rosenbrock, Ackley, Rastrigin and GaussPeaks functions in 10D on $[-100,100]^{10}$ with 500 evaluations), with the criterion, fixed in advance, of the best mean rank of the median final error (Section~\ref{sec:exp-sens}). In five and twenty dimensions the ten-dimensional settings are used in proportion ($n_0=5d$, a new map every $2d$ evaluations). For $d=30$, the CA-SAEA settings were derived from the 10D ones before any 30-dimensional run, with smaller surrogate and map sizes for computational reasons (supplementary Table~\SuppRef{tab:params}). The form and the two parameters ($m$ and $k_\ell$) of the first phase of CA-SAEA, in which several peaks are kept, were chosen in development runs on instances 0 and 1 of the MQ landscapes of Section~\ref{sec:exp-mq} and on the classic functions (supplementary Section~\SuppRef{s:onepeak}); the same values are used in all dimensions.

\emph{Development history of CA-SAEA.} Three components of CA-SAEA were changed during its development. Two were changed after runs on other test functions (a public benchmark suite for expensive optimization~\cite{chen2014cec15expensive} and the development functions above): the success rule of the trust radius was made independent of the offset of $f$, and the peaks are entered at the maximum of the interpolated surface instead of at their extrapolated locations (supplementary Section~\SuppRef{s:default}). Both were fixed before the test functions of Section~\ref{sec:exp-mq} were designed. The third change was made after the first experiments on these test functions. Until then CA-SAEA followed a single peak, and on instances 0 to 4 of the MQ landscapes it stayed on the peak that its first maps ranked highest: on the landscapes with three and five peaks in ten dimensions, \ovPkTenPFewTenCASAEA\% of its runs were on the highest peak after $10d$ evaluations and \ovPkFiftyPFewTenCASAEA\% after $50d$, whereas DYCORS, ESA and TS-DDEO reached \ovPkFiftyPFewTenDY\%, \ovPkFiftyPFewTenESA\% and \ovPkFiftyPFewTenTS\% (supplementary Section~\SuppRef{s:onepeak}). The first phase with several candidate peaks (Section~\ref{sec:casaea}) was added in response and tuned on instances 0 and 1 and on the classic functions. All MQ results reported below are therefore from instances 5 to 9, on which no algorithm was run before the algorithm was fixed; the classic functions have no further instances, and their results are not independent of this development. The test set itself, including its division into classes, was fixed after pilot runs on part of the functions and before any of these experiments. DYCORS keeps its own success rule, a threshold of $10^{-3}$ times the magnitude of its best value, as in~\cite{eriksson2019pysot}.

\emph{Implementation.} GA, DE and PSO, their controls with local search from the best point, GA+LS(surface), CA-EA, multistart Nelder--Mead, DYCORS and the initial design of CA-SAEA have two independent implementations, which agree statistically (supplementary material). CA-GA(iso), GA+LS(topo), the revised CA-SAEA and its variants have one; IPOP-CMA-ES is the reference implementation of its authors; and ESA and TS-DDEO have one re-implementation each, checked against the published results (supplementary Section~\SuppRef{sec:supp-recent}). The results of Sections~\ref{sec:exp-mq}--\ref{sec:exp-abl} were computed on a 24-thread desktop computer by a co-author who did not write the programs. The other runs were executed on a two-core virtual machine with single-threaded linear algebra. An earlier replication on a second platform of all runs then in the paper, which included the initial design of CA-SAEA, found no significant difference in any function--algorithm cell (supplementary material).

\subsection{Accuracy of the Peak Estimates}
\label{sec:exp-est}
\emph{Exact surfaces.} To test the extrapolation without interpolation error, we applied it to exact peaks of known height and location, using bands $[c_{\mathrm{lo}},c_{\mathrm{hi}}]$ at given fractions of the way from the base contour to the top (Table~\ref{tab:oracle}). For quadratic peaks the estimates are exact up to discretization: the relative height errors are at most \nOracleQuadMaxPct\% in 2, 3, 5 and 10 dimensions and the location errors below $10^{-5}$ (normalized units). For smooth non-quadratic peaks the error shrinks quickly as the band approaches the top; with the band $[0.7,0.95]$ the height error is at most \nOracleTopMaxPct\% for the Gaussian and skewed Gaussian peaks and the main peak of the Peaks function. For bands in the upper half of the peak, the second-order model~\eqref{eq:taylor2} reduces the error of the linear law $A=a_1\delta$ by factors of 3.6 to 10, as Proposition~\ref{prop:smooth} predicts for $\delta\to0$; for the lowest band, $[0.2,0.6]$, it overshoots by 20\% to 26\%, whereas the linear law undershoots by 14\% to 17\%. The non-differentiable cone, which violates the assumptions, is underestimated by 1.1\% to 2.1\%. In $n$ dimensions (ray-cast level sets, supplementary Table~\SuppRef{tab:oraclend}), the height errors stay below 3\% when the band lies close to the peak, but grow to $+23\%$ in 5D and $-38\%$ in 10D for Gaussian peaks when the evaluated points lie far below the top: the volume law amplifies non-quadratic effects as the dimension grows, so the band must be closer to the peak. The location error is $0.64\%$ of the domain width for the main peak of the Peaks function with the lowest band, $[0.2,0.6]$ (Table~\ref{tab:oracle}), and stays below $0.5\%$ in all other cases.

\begin{table*}[!t]
\centering
\caption{Contour extrapolation on exact peaks ($d=2$): relative height error $\Delta h$ in \% of $f^*-c_{\mathrm{base}}$ for the second-order (T2) and linear (lin) area models, and location error $\Delta x$ (T2, normalized units), for bands at the given fractions of the way from the base contour to the peak}
\label{tab:oracle}
\scriptsize
\begin{tabular}{lrrrrrrrrr}
\toprule
& \multicolumn{3}{c}{\thBand\ $[0.20,0.60]$} & \multicolumn{3}{c}{\thBand\ $[0.50,0.90]$} & \multicolumn{3}{c}{\thBand\ $[0.70,0.95]$} \\
\cmidrule(lr){2-4} \cmidrule(lr){5-7} \cmidrule(lr){8-10}
\thShape & $\Delta h_{\mathrm{T2}}$ & $\Delta h_{\mathrm{lin}}$ & $\Delta x$ & $\Delta h_{\mathrm{T2}}$ & $\Delta h_{\mathrm{lin}}$ & $\Delta x$ & $\Delta h_{\mathrm{T2}}$ & $\Delta h_{\mathrm{lin}}$ & $\Delta x$ \\
\midrule
\shQuad & -0.04 & 0.00 & 2.3e-16 & +0.01 & 0.00 & 1.7e-16 & 0.00 & 0.00 & 5.6e-17 \\
\shGauss & +20.03 & -16.58 & 2.0e-16 & +0.64 & -2.31 & 5.6e-17 & +0.09 & -0.67 & 1.1e-16 \\
\shSkew & +19.97 & -16.61 & 2.9e-3 & +0.64 & -2.31 & 1.8e-4 & +0.08 & -0.67 & 5.6e-5 \\
\shCone & -2.09 & -26.81 & 2.3e-16 & -1.77 & -8.31 & 0 & -1.08 & -4.25 & 2.8e-16 \\
\shPeaks & +26.30 & -13.60 & 6.4e-3 & +0.40 & -1.92 & 3.1e-4 & +0.06 & -0.56 & 8.2e-5 \\
\bottomrule
\end{tabular}

\end{table*}

\emph{Estimates from samples.} Supplementary Table~\SuppRef{tab:oneshot} compares four peak estimates computed from $N$ space-filling samples: the contour estimate (extrapolated when possible), the maximum of the same interpolated surface, a quadratic response surface fitted to the neighbors of the best sample, and the best sample itself. We report the median relative height error and the basin-hit rate, i.e., the fraction of runs in which a Nelder--Mead search started from the estimate reaches the global maximum. The contour and surface estimates have similar basin-hit rates, which shows that the interpolated surface locates the peak basin equally well in both cases. Their height estimates differ, but not consistently. Extrapolation applies in 66\% to 97\% of the 2D runs (increasing with $N$) and in 69\% to 77\% of the 5D runs; among these runs, the extrapolated height is closer to the truth than the surface maximum in 47\%, 50\%, 57\% and 52\% of the 2D runs for $N=20$, 40, 80 and 160, and in 55\% to 73\% of the 5D runs. With $N=160$ two-dimensional samples, extrapolation reduces the median height error on seven of the nine functions, by 7\% to 51\% (Peaks), but increases it 4.6-fold on Himmelblau and 2.1-fold on six-hump camel, whose peaks the interpolated surface already reproduces accurately. In 5D most samples miss the global basin, and the pooled median errors of all four estimates lie between 0.31 and 0.43. Extrapolation thus helps where the interpolated surface flattens a peak that has no data near its top, and hurts where it is already accurate.

\subsection{CA-EA: Accelerating an Evolutionary Run}
\label{sec:exp-caea}
Table~\ref{tab:ea2d} shows the two-dimensional results (convergence curves: supplementary Fig.~\SuppRef{fig:conv2d}). A GA alone reaches the target accuracy in \nGaSolvedMin\% to \nGaSolvedMax\% of the runs (never on Ackley, where its ERT is infinite), and inserting contour steps reduces its ERT by factors of \nCaGaVsGaMin\ to \nCaGaVsGaMax\ on the other eight functions. Most of this gain is due to the local search itself: GA+LS(best), which starts the same local search from the best point, is about equally fast on the five functions with at most six local optima. On the four functions with ten or more local optima (GaussPeaks, GaussPeaks21, Ackley and Rastrigin), however, CA-GA is \nCaVsLsbMultiMin\ to \nCaVsLsbMultiMax\ times faster than GA+LS(best) and succeeds in \nCaMultiSucc\ of 120 runs against \nLsbMultiSucc\ (exact McNemar test, $p=\nMcnemarP$).

The controls in the upper panel of Table~\ref{tab:ea2d} show where this advantage comes from. The two methods are not equivalent in the number of local searches: GA+LS(best) restarts only when the GA improves the best point and performs \nNlsMultiLsb\ local searches per run on these four functions, against \nNlsMultiCa\ for CA-GA. GA+LS(topo), which restarts at distinct locally best points of the archive without a contour map (\nNlsMultiTopo\ searches per run), succeeds about as often as CA-GA (\nSuccMultiTopo\ of 120 runs, $p=\nMcnemarTopoP$) but needs \nRatioMultiTopoMin\ to \nRatioMultiTopoMax\ times as many evaluations (median time to the target \nMedTimeTopo\ against \nMedTimeCa\ evaluations, paired Wilcoxon test $p=\nWilcoxTimeTopoP$). The isotropic simplex of CA-GA(iso) performs at least as well as the shape-aware one (\nSuccMultiIso\ successes, ERT ratios \nRatioMultiIsoMin\ to \nRatioMultiIsoMax\ relative to CA-GA), and starting from the surface maximum instead of the extrapolated peak changes nothing (GA+LS(surface): \nSuccMultiSurf\ successes, ratios \nRatioMultiSurfMin\ to \nRatioMultiSurfMax). The advantage of CA-GA over the classical memetic control therefore comes mainly from restarting local search on distinct peaks, and selecting these peaks with the contour tree, which ranks them by estimated height, is faster than selecting them from the evaluated points alone. In two dimensions, neither the shape of the initial simplex nor the extrapolated start point makes a measurable difference. Contour steps accelerate DE by factors of \nCaDEMin\ to \nCaDEMax\ and PSO by factors of \nCaPSOMin\ to \nCaPSOMax. On the four functions with many local optima, CA-DE and CA-PSO succeed more often than DE+LS(best) and PSO+LS(best) (\nSuccMultiDECa\ versus \nSuccMultiDELsb\ and \nSuccMultiPSOCa\ versus \nSuccMultiPSOLsb\ of 120 runs, the sums of the successful runs on these four functions in Table~\ref{tab:ea2d}); CA-PSO is also faster on all four (ERT ratios \nRatioMultiPSOLsbMin\ to \nRatioMultiPSOLsbMax), whereas DE+LS(best) is faster than CA-DE on GaussPeaks21 and Ackley (ratios \nRatioMultiDELsbMin\ to \nRatioMultiDELsbMax).

\begin{table*}[!t]
\centering
\caption{2D suite (budget 2000, 30 runs): ERT to a relative error of $10^{-6}$, with the success rate in parentheses when it is below 1; the best ERT of each row and panel is in bold. Upper panel: the GA, its four controls and CA-GA; lower panel: DE, PSO and the references. \thNls: mean number of local searches per run over the nine functions}
\label{tab:ea2d}
\scriptsize
\setlength{\tabcolsep}{3pt}
\begin{tabular}{lrrrrrrrr}
\toprule
\thFunc & GA & \shortstack{GA+\\LS(best)} & \shortstack{GA+\\LS(topo)} & \shortstack{GA+\\LS(surf.)} & \shortstack{CA-GA\\(iso)} & CA-GA & & \\
\midrule
Peaks & 4073 (0.43) & 155 & 155 & 153 & \textbf{152} & \textbf{152} & & \\
Himmelblau & 3031 (0.53) & 148 & 148 & 148 & \textbf{146} & 147 & & \\
SixHumpCamel & 1670 (0.80) & 146 & 146 & 145 & \textbf{144} & 145 & & \\
Branin & 2253 (0.67) & \textbf{147} & \textbf{147} & \textbf{147} & \textbf{147} & \textbf{147} & & \\
Rosenbrock & 18665 (0.10) & 176 & 176 & 168 & 180 & \textbf{167} & & \\
GaussPeaks & 7889 (0.23) & 952 (0.73) & 552 (0.93) & 450 (0.93) & \textbf{407 (0.97)} & 422 (0.93) & & \\
GaussPeaks21 & 19540 (0.10) & 1232 (0.67) & 920 (0.83) & \textbf{754 (0.83)} & 767 (0.83) & 764 (0.83) & & \\
Ackley & $\infty$ (0.00) & 742 (0.80) & 650 (0.83) & 423 (0.90) & \textbf{267 (0.97)} & 421 (0.90) & & \\
Rastrigin & 14342 (0.13) & 535 & 488 & 473 & 426 & \textbf{425} & & \\
\midrule
\thNls & -- & 1.2 & 6.9 & 5.2 & 5.3 & 5.3 & & \\
\midrule
\thFunc & DE & \shortstack{DE+\\LS(best)} & CA-DE & PSO & \shortstack{PSO+\\LS(best)} & CA-PSO & \shortstack{IPOP-\\CMA-ES} & MS-NM \\
\midrule
Peaks & 446 & 117 & 113 & 796 & 118 & \textbf{108} & 286 & 209 \\
Himmelblau & 634 & 111 & 110 & 735 & 110 & 108 & 207 & \textbf{47} \\
SixHumpCamel & 523 & 109 & 109 & 702 & 105 & \textbf{104} & 186 & 147 \\
Branin & 577 & 110 & 109 & 672 & 108 & 106 & 194 & \textbf{46} \\
Rosenbrock & 554 (0.97) & 124 & 128 & 1246 (0.97) & 129 & 119 & 351 & \textbf{85} \\
GaussPeaks & 1214 (0.77) & 780 (0.77) & \textbf{320 (0.97)} & 2526 (0.57) & 1303 (0.67) & 375 (0.97) & 2756 (0.50) & 958 (0.83) \\
GaussPeaks21 & 1530 (0.70) & \textbf{832 (0.77)} & 921 (0.73) & 2024 (0.67) & 1390 (0.63) & 899 (0.77) & 2794 (0.47) & 1252 (0.77) \\
Ackley & 858 & \textbf{207} & 306 & 8456 (0.23) & 509 (0.87) & 490 (0.87) & 382 & 454 \\
Rastrigin & 981 (0.97) & 709 (0.90) & 582 (0.97) & 1590 (0.87) & 697 (0.97) & \textbf{427} & 6145 (0.27) & 6791 (0.27) \\
\midrule
\thNls & -- & 1.2 & 5.2 & -- & 1.2 & 5.5 & -- & -- \\
\bottomrule
\end{tabular}

\end{table*}

Supplementary Table~\SuppRef{tab:eand} reports the five- and ten-dimensional results. On Rosenbrock and Ackley the GA alone never reaches the target, whereas CA-GA succeeds in \nNdVRosenbrockCa\% and \nNdVAckleyCa\% of the runs in 5D and in \nNdOZRosenbrockCa\% and \nNdOZAckleyCa\% in 10D (GA+LS(best): \nNdVRosenbrockLsb\%, \nNdVAckleyLsb\%, \nNdOZRosenbrockLsb\% and \nNdOZAckleyLsb\%). On the rugged Rastrigin function the contour step is less effective than starting from the best point in 5D, and in 10D every variant with local search ends with a larger median error than the GA alone, because local searches converge to the nearest of the many local optima. GaussPeaks with 21 peaks remains hard for all methods in 5D and 10D.

\emph{Extrapolated peak versus surface maximum.} Within CA-EA, starting the local search from the extrapolated peak (CA-GA) and from the maximum of the interpolated surface in the same contour region (GA+LS(surface)) gives practically the same results (supplementary Fig.~\SuppRef{fig:conv2d}). Over the 17 functions of the 2D, 5D and 10D suites, the paired Wilcoxon tests on the times to the target yield a minimum Holm-adjusted $p$-value of 0.26. Nelder--Mead corrects a start point error of the size we observe within a few iterations, so the closed-loop benefit of a more accurate start is small.

\subsection{CA-SAEA on Multi-Peak Test Landscapes}
\label{sec:exp-mq}
\emph{Choice of the test functions.} No set of test functions represents all optimization problems, and a benchmark assembled for a competition covers the cases its organizers selected. We therefore test CA-SAEA on functions that we designed and chose ourselves, and we state why. The idea of this paper is that the contour map of the evaluated solutions reveals the peaks of a landscape, so that each peak can be remembered and entered where it is highest. The main test functions are accordingly landscapes made of one or a few smooth peaks that differ in height and orientation, with different elongations from landscape to landscape (class~A); this is the structure the method is built for, and it is common in practice, where a small number of distinct good designs compete. To show where the scope of the method ends, we add classic rugged functions with many deep local peaks (class~B) and landscapes with ten and thirty peaks; their results are reported in the same way as those of class~A.

\emph{Multi-peak quadratic landscapes (MQ).} For a dimension $d$, a number of peaks $K$ and a condition number $\kappa$, the function to be maximized on $[-5,5]^d$ is the upper envelope of $K$ concave quadratic peaks,
\begin{equation}
f(\mathbf{x})=\max_{1\le k\le K}\Bigl[-b_k-(\mathbf{x}-\mathbf{c}_k)^{\top}\mathbf{R}_k^{\top}\boldsymbol{\Lambda}\,\mathbf{R}_k(\mathbf{x}-\mathbf{c}_k)\Bigr],
\label{eq:mq}
\end{equation}
with $\boldsymbol{\Lambda}=\mathrm{diag}(\lambda_1,\dots,\lambda_d)$ and $\lambda_i=1+(\kappa-1)\bigl((i-1)/(d-1)\bigr)^2$, so that the longest axis of a peak is $\sqrt{\kappa}$ times its shortest. The centres $\mathbf{c}_k$ are drawn uniformly in $[-4,4]^d$ at a mutual distance of at least $1.2\sqrt{d}$, and each $\mathbf{R}_k$ is an independent random orthogonal matrix. The first peak is the highest, $b_1=0$; for $k\ge 2$, $b_k=\rho_k\min_{j\ne k}(\mathbf{c}_k-\mathbf{c}_j)^{\top}\mathbf{R}_j^{\top}\boldsymbol{\Lambda}\mathbf{R}_j(\mathbf{c}_k-\mathbf{c}_j)$ with $\rho_k$ uniform in $[0.1,0.5]$. This choice makes every $\mathbf{c}_k$ a strict local maximizer and $\mathbf{c}_1$ the global one, with $f(\mathbf{c}_1)=0$, so the error of a solution is $-f(\mathbf{x})$. An instance is determined by $d$, $K$, $\kappa$ and an instance number.

\emph{The test set.} Class~A consists of the MQ landscapes with $K\in\{1,3,5\}$ and $\kappa\in\{10^2,10^4,10^6\}$, five instances each (instances 5 to 9; instances 0 to 4 served the development, Section~\ref{sec:exp-setup}), and of the classic functions Sphere, Rosenbrock and Ackley, whose landscapes are dominated by a single peak. Class~B consists of the classic Schwefel and Rastrigin functions. The MQ landscapes with $K=10$ (same $\kappa$, five instances each) and with $K=30$ ($\kappa\in\{1,10^2\}$, five instances each) are reported separately, and so is GaussPeaks, the 21-peak function of the suites above (a new random landscape in every run). All functions are used in five and ten dimensions with $50d$ evaluations and 20 runs; instance 5 of each MQ configuration with $K\le10$ and the five classic functions are also run in twenty and thirty dimensions (Section~\ref{sec:exp-dim}). CA-SAEA is compared with DYCORS, ESA and TS-DDEO at the same numbers of evaluations. Two algorithms are called different on a function when the rank-sum test (the signed-rank test between CA-SAEA and its variants, which share the initial design) rejects equality at the 0.05 level and their median errors differ by more than 1\%, a threshold fixed before these experiments. Of two algorithms that differ, the better one is the one that the ranks of the test favor; where the runs of both end on different peaks, this is not always the one with the lower median (two functions in ten dimensions, named in the captions of supplementary Tables~\SuppRef{tab:mqA} and~\SuppRef{tab:mqother}). The settings of all algorithms are those of Section~\ref{sec:exp-setup}.

\emph{Results at the full budget.} Table~\ref{tab:mq} gives, for each class, the number of functions on which CA-SAEA is better than, not different from and worse than each of the three algorithms after $50d$ evaluations; supplementary Tables~\SuppRef{tab:mqA} and~\SuppRef{tab:mqother} list the median errors on every function. On class~A, counting both dimensions, CA-SAEA is better than DYCORS on \nvALowDYW\ of the \nvALowDYN\ functions, not different on \nvALowDYT\ and worse on \nvALowDYL; against TS-DDEO it is better on \nvALowTSW, not different on \nvALowTST\ and worse on \nvALowTSL. The advantage over DYCORS is larger in ten dimensions than in five and includes Sphere, Rosenbrock and Ackley in both. DYCORS is the closest relative of CA-SAEA: both choose every evaluation with an RBF surrogate among perturbations of a center, and CA-SAEA adds the contour map, a tracker with its own trust radius for every peak, and the global steps. ESA is better than CA-SAEA on class~A: CA-SAEA is worse on \nvALowESAL\ of the \nvALowESAN\ functions, not different on \nvALowESAT\ and better on \nvALowESAW. ESA reaches a higher accuracy on these smooth peaks, also on the landscapes with a single peak, where convergence is the only task.

On class~B, CA-SAEA is not worse than any of the three algorithms on either function in either dimension: it does not differ from DYCORS, it is better than ESA on Rastrigin in five dimensions and on Schwefel in ten, and it is better than TS-DDEO in all four cases. These rugged functions violate the assumption of a few smooth peaks, and the errors of all four algorithms remain large.

With ten peaks, the counts over both dimensions are \nvKTenLowDYWTL\ against DYCORS, \nvKTenLowESAWTL\ against ESA and \nvKTenLowTSWTL\ against TS-DDEO, and with thirty peaks \nvKThirtyLowDYWTL, \nvKThirtyLowESAWTL\ and \nvKThirtyLowTSWTL; on GaussPeaks CA-SAEA is better than TS-DDEO and does not differ from the other two. The advantage of CA-SAEA over DYCORS and TS-DDEO thus becomes smaller as the number of peaks grows: with thirty peaks it differs from each of the three algorithms on a minority of the landscapes. None of the three algorithms is better than CA-SAEA on both class~A and class~B.

\begin{figure}[!t]
\centering
\includegraphics[width=\columnwidth]{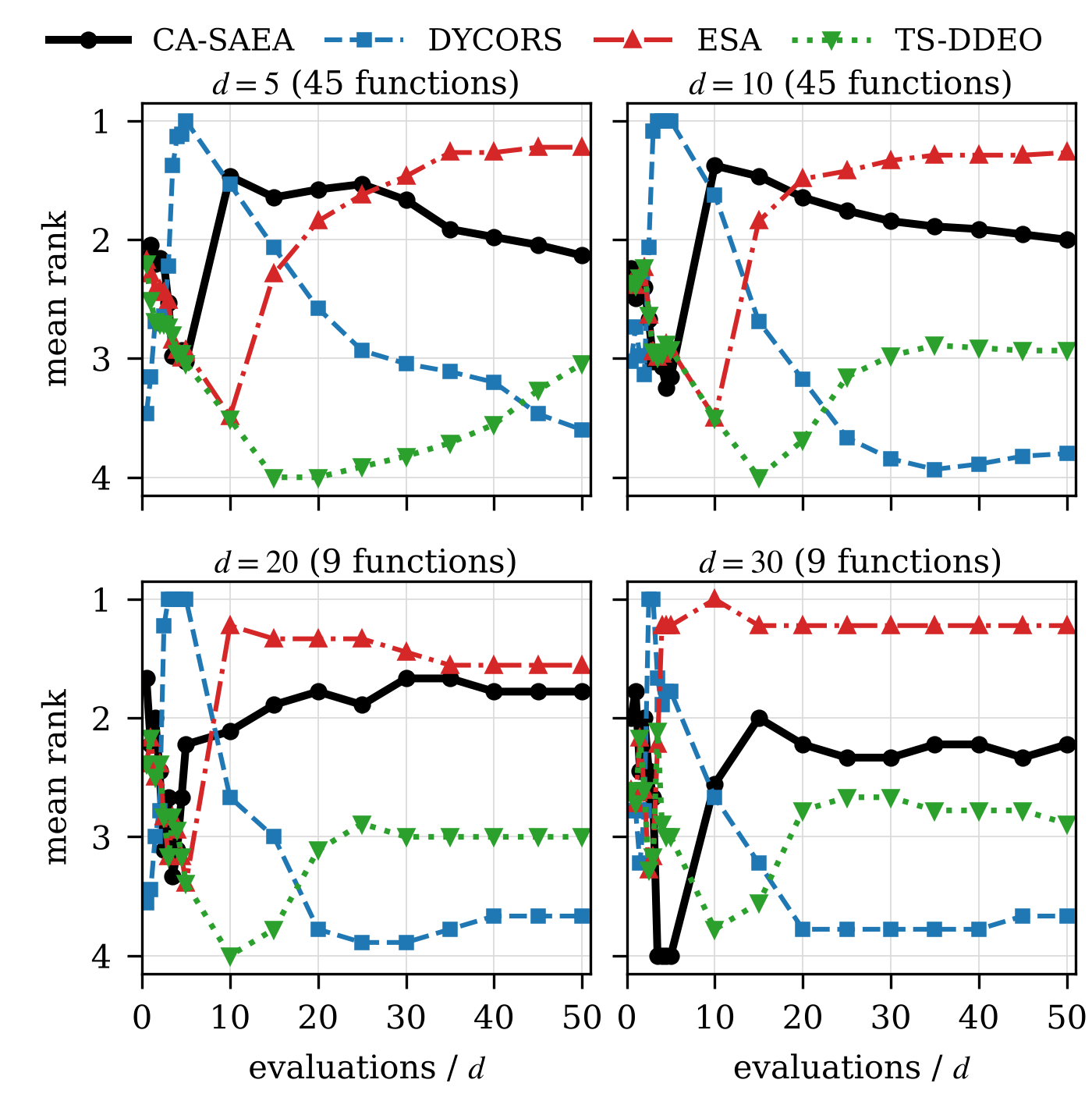}
\caption{Class-A MQ landscapes: mean rank of the four algorithms by their median error over the number of evaluations (rank 1 is best)}
\label{fig:mqrank}
\end{figure}

\emph{Course of the search.} Fig.~\ref{fig:mqrank} shows the mean rank of the four algorithms on the class-A MQ landscapes over the budget. Once its initial design of $5d$ points is evaluated, CA-SAEA improves fastest: it has the best mean rank from $\nvLeadFromFive d$ to $\nvLeadToFive d$ evaluations in five dimensions and from $\nvLeadFromTen d$ to $\nvLeadToTen d$ in ten (ESA and TS-DDEO spend their first 50 or 100 evaluations on the initial samples). ESA then overtakes it, and at $50d$ evaluations the mean ranks are \nvMRankESAFive\ for ESA, \nvMRankCASAEAFive\ for CA-SAEA, \nvMRankTSFive\ for TS-DDEO and \nvMRankDYFive\ for DYCORS in five dimensions, and \nvMRankESATen, \nvMRankCASAEATen, \nvMRankTSTen\ and \nvMRankDYTen\ in ten. Supplementary Table~\SuppRef{tab:budget} gives the counts of Table~\ref{tab:mq} after $10d$ to $50d$ evaluations.

\begin{figure}[!t]
\centering
\includegraphics[width=\columnwidth]{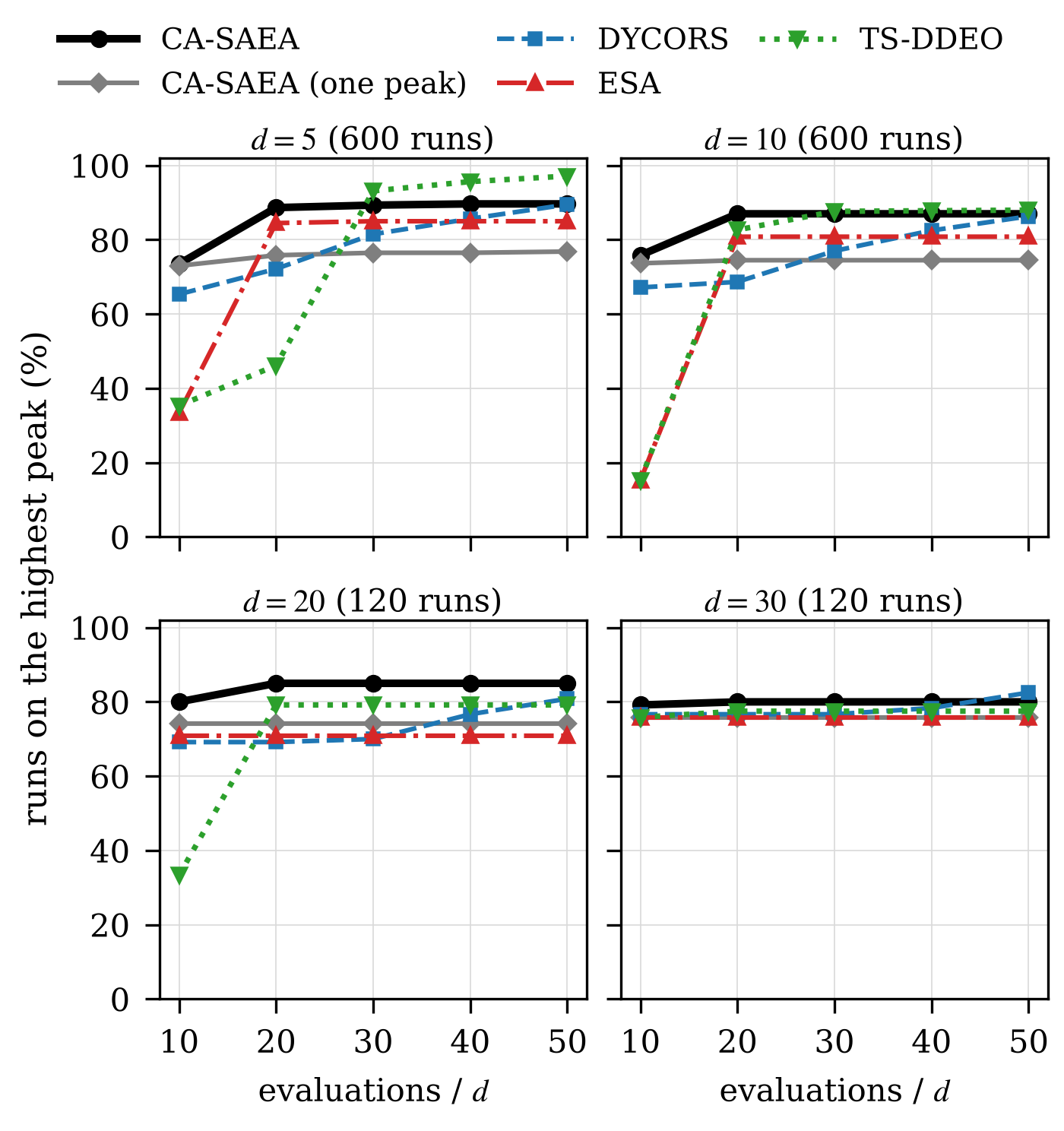}
\caption{MQ landscapes with three and five peaks: share of the runs on the highest peak (error below $\min_{k\ge2}b_k$) over the number of evaluations}
\label{fig:mqpeak}
\end{figure}

\emph{Keeping several peaks.} On an MQ landscape, a run whose error is below $\min_{k\ge2}b_k$ has its best solution on the highest peak and above the top of every other peak. CA-SAEA (one peak), the version without the first phase, has \nvPkTenPFewFiveOne\% of its runs in five dimensions and \nvPkTenPFewTenOne\% in ten on the highest peak of the class-A landscapes with three and five peaks after $10d$ evaluations, and \nvPkFiftyPFewFiveOne\% and \nvPkFiftyPFewTenOne\% after $50d$: it stays on the peak that its first maps rank highest. Keeping several peaks first changes the shares after $50d$ evaluations to \nvPkFiftyPFewFiveCASAEA\% and \nvPkFiftyPFewTenCASAEA\%, and on the landscapes with ten peaks from \nvPkFiftyKTenFiveOne\% and \nvPkFiftyKTenTenOne\% to \nvPkFiftyKTenFiveCASAEA\% and \nvPkFiftyKTenTenCASAEA\% (supplementary Table~\SuppRef{tab:peak}; with three and five peaks, DYCORS, ESA and TS-DDEO reach between \nvPkFiftyPOppMinLow\% and \nvPkFiftyPOppMaxLow\%). Table~\ref{tab:keep} compares the final errors of the two versions: over five and ten dimensions CA-SAEA is better on \nvKeepAllLowOneW\ of the \nvKeepAllLowOneN\ functions and worse on \nvKeepAllLowOneL. On the landscapes with three and five peaks the counts are \nvKeepPFewLowOneWTL, on those with ten peaks \nvKeepKTenLowOneWTL, and on those with a single peak, where the first phase has no choice to make, \nvKeepKOneLowOneWTL. The first phase thus raises the share of the runs on the highest peak to the level that the other three algorithms reach, and on the landscapes with a single peak the two versions do not differ significantly on any function (Fig.~\ref{fig:mqpeak}).

\begin{table}[!t]
\centering
\caption{CA-SAEA against DYCORS, ESA and TS-DDEO at $50d$ evaluations: number of functions on which CA-SAEA is better / not different / worse (two-sided rank-sum test, $\alpha=0.05$, 20 runs; medians within 1\% count as not different)}
\label{tab:mq}
\scriptsize
\setlength{\tabcolsep}{2.4pt}
\begin{tabular}{lccccccc}
\toprule
 & & \multicolumn{3}{c}{$d=5$} & \multicolumn{3}{c}{$d=10$}\\
Class & $n$ & DYCORS & ESA & TS-DDEO & DYCORS & ESA & TS-DDEO\\
\midrule
A: MQ, $K\le 5$ & 45 & 34/11/0 & 1/15/29 & 29/13/3 & 43/2/0 & 0/19/26 & 31/14/0 \\
A: classic & 3 & 3/0/0 & 0/2/1 & 2/1/0 & 3/0/0 & 0/1/2 & 2/0/1 \\
B: rugged & 2 & 0/2/0 & 1/1/0 & 2/0/0 & 0/2/0 & 1/1/0 & 2/0/0 \\
MQ, $K=10$ & 15 & 4/8/3 & 0/12/3 & 4/11/0 & 12/3/0 & 1/11/3 & 6/9/0 \\
MQ, $K=30$ & 10 & 0/10/0 & 0/10/0 & 6/3/1 & 3/7/0 & 0/9/1 & 0/9/1 \\
GaussPeaks & 1 & 0/1/0 & 0/1/0 & 1/0/0 & 0/1/0 & 0/1/0 & 1/0/0 \\
\midrule
All & 76 & 41/32/3 & 2/41/33 & 44/28/4 & 61/15/0 & 2/42/32 & 42/32/2 \\
\bottomrule
\end{tabular}

\end{table}

\begin{table}[!t]
\centering
\caption{CA-SAEA against CA-SAEA (one peak), the version without the first phase, at $50d$ evaluations: number of functions on which CA-SAEA is better / not different / worse (two-sided signed-rank test on the 20 paired runs, $\alpha=0.05$; medians within 1\% count as not different). The column $n$ gives the numbers of functions in the four dimensions}
\label{tab:keep}
\scriptsize
\setlength{\tabcolsep}{3pt}
\begin{tabular}{lccccc}
\toprule
 & & \multicolumn{1}{c}{$d=5$} & \multicolumn{1}{c}{$d=10$} & \multicolumn{1}{c}{$d=20$} & \multicolumn{1}{c}{$d=30$}\\
Class & $n$ & W/T/L & W/T/L & W/T/L & W/T/L\\
\midrule
A: MQ, $K\le 5$ & 45/45/9/9 & 2/43/0 & 3/41/1 & 2/7/0 & 1/8/0 \\
A: classic & 3 & 0/3/0 & 0/3/0 & 0/3/0 & 0/3/0 \\
B: rugged & 2 & 0/2/0 & 0/2/0 & 0/2/0 & 0/2/0 \\
MQ, $K=10$ & 15/15/3/3 & 4/11/0 & 4/11/0 & 0/3/0 & 1/2/0 \\
MQ, $K=30$ & 10/10/0/0 & 1/9/0 & 4/6/0 & -- & -- \\
GaussPeaks & 1/1/0/0 & 0/1/0 & 0/1/0 & -- & -- \\
\midrule
All & 76/76/17/17 & 7/69/0 & 11/64/1 & 2/15/0 & 2/15/0 \\
\bottomrule
\end{tabular}

\end{table}

\subsection{Dependence on the Dimension}
\label{sec:exp-dim}
In twenty and thirty dimensions the test set consists of instance~5 of every MQ configuration with $K\le10$ and of the five classic functions, \nvAllTwentyDYN\ functions per dimension. Table~\ref{tab:dims} gives the counts of Table~\ref{tab:mq} for these dimensions, and supplementary Table~\SuppRef{tab:mqhigh} the median errors. On the \nvATwentyDYN\ functions of class~A, CA-SAEA is better than DYCORS on \nvATwentyDYW\ and worse on \nvATwentyDYL\ in twenty dimensions, and better on \nvAThirtyDYW\ and worse on \nvAThirtyDYL\ in thirty. Against TS-DDEO the counts are \nvATwentyTSWTL\ and \nvAThirtyTSWTL, and against ESA \nvATwentyESAWTL\ and \nvAThirtyESAWTL. On the two rugged functions of class~B the counts are \nvBTwentyDYWTL\ and \nvBThirtyDYWTL\ against DYCORS, \nvBTwentyESAWTL\ and \nvBThirtyESAWTL\ against ESA, and \nvBTwentyTSWTL\ and \nvBThirtyTSWTL\ against TS-DDEO; with ten peaks they are \nvKTenHighDYWTL, \nvKTenHighESAWTL\ and \nvKTenHighTSWTL\ over the two dimensions. The comparison with DYCORS on class~A thus changes little with the dimension: in five and ten dimensions CA-SAEA was better on \nvAFiveDYW\ and \nvATenDYW\ of \nvAFiveDYN\ functions, and it is worse on none in any of the four dimensions. Against TS-DDEO it is better on about two thirds of the class-A functions up to twenty dimensions and on a third in thirty, where half of the functions show no difference, and ESA is better on most class-A functions in every dimension. On class~B, CA-SAEA is not worse than any of the three algorithms up to twenty dimensions; in thirty, DYCORS is better on both functions and ESA on Rastrigin. Over the budget, ESA has the best mean rank on the class-A MQ landscapes of these dimensions from $10d$ evaluations on, and CA-SAEA the second best (Fig.~\ref{fig:mqrank}); at $50d$ evaluations the mean ranks of ESA, CA-SAEA, TS-DDEO and DYCORS are \nvMRankESATwenty, \nvMRankCASAEATwenty, \nvMRankTSTwenty\ and \nvMRankDYTwenty\ in twenty dimensions and \nvMRankESAThirty, \nvMRankCASAEAThirty, \nvMRankTSThirty\ and \nvMRankDYThirty\ in thirty.

The first phase is used unchanged in these dimensions. After $50d$ evaluations, \nvPkFiftyPFewTwentyOne\% of the runs of CA-SAEA (one peak) and \nvPkFiftyPFewTwentyCASAEA\% of those of CA-SAEA are on the highest peak of the landscapes with three and five peaks in twenty dimensions, and \nvPkFiftyPFewThirtyOne\% and \nvPkFiftyPFewThirtyCASAEA\% in thirty; with ten peaks the shares are \nvPkFiftyKTenTwentyOne\% and \nvPkFiftyKTenTwentyCASAEA\%, and \nvPkFiftyKTenThirtyOne\% and \nvPkFiftyKTenThirtyCASAEA\%. In the final error, CA-SAEA is better than CA-SAEA (one peak) on \nvKeepAllHighOneW\ of the \nvKeepAllHighOneN\ functions of the two dimensions and worse on \nvKeepAllHighOneL\ (Table~\ref{tab:keep}). The first phase thus acts in twenty and thirty dimensions as it does in five and ten: more runs end on the highest peak, and the final error is worse on no function (on one of \nvKeepAllLowOneN\ in five and ten dimensions). With three and five peaks, the other three algorithms have \nvPkFiftyPOppMinHigh\% to \nvPkFiftyPOppMaxHigh\% of their runs on the highest peak in these dimensions.

\begin{table}[!t]
\centering
\caption{CA-SAEA in twenty and thirty dimensions ($50d$ evaluations, 20 runs): number of functions on which CA-SAEA is better / not different / worse}
\label{tab:dims}
\scriptsize
\setlength{\tabcolsep}{2.4pt}
\begin{tabular}{lccccccc}
\toprule
 & & \multicolumn{3}{c}{$d=20$} & \multicolumn{3}{c}{$d=30$}\\
Class & $n$ & DYCORS & ESA & TS-DDEO & DYCORS & ESA & TS-DDEO\\
\midrule
A: MQ, $K\le 5$ & 9 & 9/0/0 & 1/2/6 & 7/2/0 & 7/2/0 & 0/2/7 & 4/5/0 \\
A: classic & 3 & 2/1/0 & 0/0/3 & 1/0/2 & 3/0/0 & 0/0/3 & 0/1/2 \\
B: rugged & 2 & 0/2/0 & 1/1/0 & 2/0/0 & 0/0/2 & 1/0/1 & 2/0/0 \\
MQ, $K=10$ & 3 & 3/0/0 & 0/0/3 & 2/1/0 & 0/3/0 & 0/2/1 & 0/3/0 \\
\midrule
All & 17 & 14/3/0 & 2/3/12 & 12/3/2 & 10/5/2 & 1/4/12 & 6/9/2 \\
\bottomrule
\end{tabular}

\end{table}

\subsection{Ablation of CA-SAEA}
\label{sec:exp-abl}
Table~\ref{tab:abl} compares the default CA-SAEA with two variants that start from the same initial designs. CA-SAEA (best) draws no contour map: all its local steps sample around the best solution, with a single trust radius. Over the \nvAblAllLowBestN\ functions of the five- and ten-dimensional test set, the default is better than this variant on \nvAblAllLowBestW, not different on \nvAblAllLowBestT\ and worse on \nvAblAllLowBestL; on class~A the counts are \nvAblALowBestWTL. The default differs from this variant by the contour map, with a tracker and a trust radius for every peak, and by the additional global steps that it takes when all its peaks are explored (Section~\ref{sec:casaea}); it is thus better on more than half of the functions and worse on none, and the counts are \nvAblAllFiveBestWTL\ in five dimensions and \nvAblAllTenBestWTL\ in ten. CA-SAEA (extrap.) enters every peak at the location extrapolated by the area law instead of at the maximum of the interpolated surface. The default is better than it on \nvAblAllLowExtrapW\ functions, not different on \nvAblAllLowExtrapT\ and worse on \nvAblAllLowExtrapL\ (\nvAblALowExtrapWTL\ on class~A). Neither entry point is thus better than the other overall, and the default uses the simpler one. The comparison with the third variant, CA-SAEA (one peak), is given in Section~\ref{sec:exp-mq}. Supplementary Section~\SuppRef{s:ablation} lists the results on every function.

\begin{table}[!t]
\centering
\caption{Ablation: the default CA-SAEA against CA-SAEA (extrap.), which enters the peaks at their extrapolated locations, and CA-SAEA (best), which draws no contour map ($50d$ evaluations, 20 paired runs, signed-rank test): number of functions on which the default is better / not different / worse}
\label{tab:abl}
\scriptsize
\setlength{\tabcolsep}{3pt}
\begin{tabular}{lccccc}
\toprule
 & & \multicolumn{2}{c}{$d=5$} & \multicolumn{2}{c}{$d=10$}\\
Class & $n$ & extrap. & best & extrap. & best\\
\midrule
A: MQ, $K\le 5$ & 45 & 0/45/0 & 17/28/0 & 1/41/3 & 36/9/0 \\
A: classic & 3 & 0/3/0 & 2/1/0 & 0/3/0 & 3/0/0 \\
B: rugged & 2 & 0/2/0 & 1/1/0 & 1/1/0 & 0/2/0 \\
MQ, $K=10$ & 15 & 0/15/0 & 5/10/0 & 1/13/1 & 10/5/0 \\
MQ, $K=30$ & 10 & 0/10/0 & 3/7/0 & 0/10/0 & 7/3/0 \\
GaussPeaks & 1 & 0/1/0 & 0/1/0 & 0/1/0 & 0/1/0 \\
\midrule
All & 76 & 0/76/0 & 28/48/0 & 3/69/4 & 56/20/0 \\
\bottomrule
\end{tabular}

\end{table}

\subsection{Sensitivity to the Configuration}
\label{sec:exp-sens}
Supplementary Table~\SuppRef{tab:sens} shows the development-suite results from which the settings of CA-SAEA for the budget of $50d$ evaluations were chosen (these runs were made with the initial design, that is, CA-SAEA (extrap.) with the earlier success rule and without the first phase). The two configurations with $n_0=5d$ had a lower median error than DYCORS on four of the five functions and those with $n_0=2(d+1)$ on three, and the chosen configuration ($n_0=5d$, $L=4$) ranked first. The four configurations differ little except on Ackley, where their median errors span a factor of 67; DYCORS is better than all of them on GaussPeaks.

\subsection{Computational Cost}
\label{sec:exp-cost}
A contour step on at most 600 points takes 0.35 to 1.5~s in our first implementation (2-D map or 10-D rays). In the runs of Sections~\ref{sec:exp-mq} and~\ref{sec:exp-dim}, a ten-dimensional run of CA-SAEA (500 evaluations, \nvMapsTen\ maps, one RBF fit per evaluation) took a median of \nvTimeCASAEATen~s, against \nvTimeDYTen~s for DYCORS, \nvTimeESATen~s for ESA and \nvTimeTSTen~s for TS-DDEO; a thirty-dimensional run (1500 evaluations, a map with 256 rays every 50 evaluations) took \nvTimeCASAEAThirty~s, against \nvTimeDYThirty~s, \nvTimeESAThirty~s and \nvTimeTSThirty~s (wall-clock times with 24 runs in parallel on a 24-thread processor). In these two dimensions CA-SAEA is the slowest of the four per run. Its overhead is negligible when an evaluation takes more than a few seconds, the regime for which it is designed.

\section{Discussion}
\label{sec:discussion}

\emph{What the contour map contributes.} The experiments separate three contributions of the contour map. (i) \emph{Topology}: the persistence-pruned contour tree decomposes the evaluated region into peaks, each with its own basin, trusted contours and, in CA-SAEA, its own trust radius and memory. Together with the peak memory, this decomposition makes CA-EA restart local search on distinct peaks, which gives it its advantage over local search from the best point on functions with many local optima. In CA-SAEA, the variant that draws no map is worse than the default on more than half of the test functions and better on none (Section~\ref{sec:exp-abl}). (ii) \emph{Shape}: the second-moment matrix of the innermost trusted contour orients the initial simplex of Nelder--Mead in CA-EA and the sampling distribution in CA-SAEA. For Nelder--Mead, which adapts its simplex anyway, an isotropic simplex of the same size works at least as well in two dimensions; in CA-SAEA the ablation measures the contribution of the map as a whole and does not isolate that of the shape. (iii) \emph{Extrapolation}: the area law turns the nested contours into an estimate of the peak height and location. It is exact on quadratic peaks, accurate near the top of smooth ones (Table~\ref{tab:oracle}) and, with enough samples, improves the estimated heights on most two-dimensional functions (supplementary Table~\SuppRef{tab:oneshot}); as a start point for local search, the simpler maximum of the interpolated surface in the same region gave results that did not differ significantly on almost all functions (Sections~\ref{sec:exp-caea} and~\ref{sec:exp-abl}), which is why CA-SAEA enters its peaks at the surface maximum.

\emph{Early maps and the choice of a peak.} A contour map is only as reliable as the solutions it is drawn from. The first maps of an expensive run are drawn from a few tens of solutions, which is when the probability of ranking the peaks wrongly is highest. A version of CA-SAEA that relied on the highest-ranked peak from its first map on stayed on that peak for the rest of the run (Section~\ref{sec:exp-mq}). We attribute this to a choice that reinforces itself: the peak that receives the local steps improves and keeps its rank, while the other peaks receive none and cannot overtake it. Keeping several peaks during the first maps and giving each of them evaluations lets a peak that looks small on an early map grow on the later ones. The number of peaks kept is then reduced step by step, because evaluations divided among several peaks until the end cost accuracy (supplementary Section~\SuppRef{s:onepeak}). We expect the same to hold for other uses of a map built during the search: a decision taken from an early map should stay open until the map has been redrawn from solutions that could contradict it.

\emph{Low and high dimensions.} From five to thirty dimensions, CA-SAEA is better than DYCORS on \nvALowDYW\ of the \nvALowDYN\ class-A functions in five and ten dimensions and on \nvAHighDYW\ of the \nvAHighDYN\ in twenty and thirty, and worse on none; within the range tested, this advantage does not depend on the dimension. In thirty dimensions, however, the advantage over TS-DDEO is limited to a third of the class-A functions, and DYCORS is better on the two rugged functions of class~B (Section~\ref{sec:exp-dim}). A contour map describes the landscape in a low-dimensional space, so in a high-dimensional problem the question becomes which dimensions the map should be computed in and how the dimension should be reduced. We regard these two choices as the main open problem of the approach.

\emph{Limitations.} The method assumes a few peaks that are smooth on the scale of the evaluated points. On rugged landscapes such as Rastrigin and Schwefel, the interpolated surface smooths out the fine structure; the peaks of the map then correspond to the envelope of many local optima, and local searches end in nearby local optima. On Rastrigin in ten dimensions the GA alone reached a smaller median final error than every variant with local search (Section~\ref{sec:exp-caea}), and CA-SAEA has no advantage over DYCORS on these functions. Its advantage over DYCORS and TS-DDEO also becomes smaller as the number of peaks grows, and with thirty peaks little of it is left (Section~\ref{sec:exp-mq}). On the smooth peaks themselves, the recent ESA reaches a higher accuracy than CA-SAEA at the full budget, also on landscapes with a single peak, where convergence is the only task; the convergence of CA-SAEA on a peak, once the peak is chosen, is therefore what should be improved first. The ray-cast level sets assume regions that are star-shaped around the peak and need the number of rays to grow with the dimension. Their cost, together with the $O(N^3)$ RBF fit, made us reduce the number of rays and the map frequency for $d=30$, and dimensions well beyond 30 would require different level-set descriptors. The rays see only the part of a region that is visible from the peak and miss the long axes of an ill-conditioned peak, and the peak memory uses a fixed radius $r_e$. Finally, the overhead of a contour step pays off only when evaluations are expensive. In practice the method is meant for landscapes with a few peaks that are smooth to moderately rugged on the scale of the evaluated points, evaluations costly enough that the map's overhead is negligible, and at most a few tens of dimensions.

\section{Conclusion}
\label{sec:conclusion}

This paper proposed contour-assisted evolutionary computation (CA-EC), in which an evolutionary algorithm reads its own evaluated solutions as a topographic survey and learns the peak structure of the landscape from them. A contour map restricted to the mapped area decomposes the evaluated region into peaks, with a persistence-pruned contour tree in two dimensions and with hill climbing on the surface and ray-cast level sets in higher dimensions. From the nested contours of each peak, the area law estimates the height, location and shape of a summit that no evaluated solution has reached, and a memory of the explored peaks directs the search to new ones. Two algorithms use the map. CA-EA starts a local search from the extrapolated summit. CA-SAEA, for expensive problems, keeps a trust radius and a memory for every peak and enters each peak at the maximum of the interpolated surface; it first keeps several peaks of its early maps and then concentrates on one.

In two dimensions, CA-EA reduces the number of evaluations required by GA, DE and PSO substantially; most of this gain comes from the local search itself, and on functions with many local optima CA-EA reaches the target accuracy in more runs than local search from the best point. With $50d$ evaluations on landscapes with one to five smooth peaks and on classic functions dominated by one peak, CA-SAEA is better than the well-established DYCORS on \nvALowDYW\ of \nvALowDYN\ functions in five and ten dimensions and worse on none, and better than the recent TS-DDEO on \nvALowTSW\ and worse on \nvALowTSL; in twenty and thirty dimensions the counts are \nvAHighDYWTL\ against DYCORS and \nvAHighTSWTL\ against TS-DDEO. On two classic rugged functions it is not worse than any of the three algorithms compared in five, ten and twenty dimensions. Keeping several peaks first brings as many runs to the highest peak as the other algorithms reach, whereas a version that follows a single peak from its first map on stays on it. Entering the peaks at the maximum of the surface instead of at their extrapolated locations gives results that do not differ significantly on \nvAblAllLowExtrapT\ of the \nvAblAllLowExtrapN\ functions.

The method has clear limits, which we leave to future work. The recent ESA reaches a higher accuracy than CA-SAEA on smooth peaks at the full budget, so the convergence on a peak, once it is chosen, can be improved. The advantage over DYCORS and TS-DDEO becomes smaller as the number of peaks grows. In thirty dimensions DYCORS is better on the two rugged functions of class~B, and over twenty and thirty dimensions ESA is better on \nvAHighESAL\ of the \nvAHighESAN\ functions of class~A. A contour map is by nature a low-dimensional description: which dimensions to compute the map in and how to reduce the dimension of a high-dimensional problem are the most important questions left open by this paper. The implementations, the generator of the test landscapes and all raw results will be released publicly upon publication.

\section*{Acknowledgment}
The original idea of this work was proposed by Professor ZHANG Jun. Under Professor Zhang's direction, an AI system, Claude (Anthropic), was used to implement the algorithms and the experiments, and help complete the manuscript. The other authors and students reviewed and proofread the code, the results and the text. The authors take full responsibility for the content.

\end{document}